\documentclass{article}

    \PassOptionsToPackage{numbers, compress}{natbib}

    \usepackage[final]{neurips_2024}

\usepackage[utf8]{inputenc} 
\usepackage[T1]{fontenc}    
\usepackage{hyperref}       
\usepackage{url}            
\usepackage{booktabs}       
\usepackage{amsfonts}       
\usepackage{nicefrac}       
\usepackage{microtype}      
\usepackage{xcolor}         

\usepackage{amsmath}
\usepackage{booktabs}
\usepackage{algorithm}

\usepackage{multirow}
\usepackage[normalem]{ulem}
\usepackage{graphicx}
\usepackage{subfigure}
\usepackage{amssymb}
\usepackage{wrapfig}

\usepackage{algpseudocode}
\usepackage{adjustbox}
\usepackage{amsthm}

\usepackage{hyperref}

\newtheorem{theorem}{Theorem}
\newtheorem{lemma}{Lemma}

\title{Bridging Gaps: Federated Multi-View Clustering in Heterogeneous Hybrid Views}

\author{%
  Xinyue Chen$^{1}$\quad Yazhou Ren$^{1,2,}$\thanks{Corresponding Author (yazhou.ren@uestc.edu.cn).} \quad Jie Xu$^{1}$\quad \\ \textbf{Fangfei Lin$^{1}$\quad  Xiaorong Pu$^{1,2}$\quad Yang Yang$^{1}$} \\
    $^{1}$School of Computer Science and Engineering, University of Electronic Science \\ and Technology of China, China; 
    $^{2}$Shenzhen Institute for Advanced Study, \\ University of Electronic Science and Technology of China, China
}

\begin{document}

\maketitle

\begin{abstract}
Recently, federated multi-view clustering (FedMVC) has emerged to explore cluster structures in multi-view data distributed on multiple clients.
Existing approaches often assume that clients are isomorphic and all of them belong to either single-view clients or multi-view clients.
Despite their success, these methods also present limitations when dealing with practical FedMVC scenarios involving heterogeneous hybrid views, where a mixture of both single-view and multi-view clients exhibit varying degrees of heterogeneity.
In this paper, we propose a novel FedMVC framework, which concurrently addresses two challenges associated with heterogeneous hybrid views, i.e., client gap and view gap.
To address the client gap, we design a local-synergistic contrastive learning approach that helps single-view clients and multi-view clients achieve consistency for mitigating heterogeneity among all clients.
To address the view gap, we develop a global-specific weighting aggregation method, which encourages global models to learn complementary features from hybrid views.
The interplay between local-synergistic contrastive learning and global-specific weighting aggregation mutually enhances the exploration of the data cluster structures distributed on multiple clients.
Theoretical analysis and extensive experiments demonstrate that our method can handle the heterogeneous hybrid views in FedMVC and outperforms state-of-the-art methods.
The code is available at \url{https://github.com/5Martina5/FMCSC}.
\end{abstract}

\section{Introduction}

Recent advancements in sensors and the internet have allowed many distributed clients to collect unlabeled data from multiple views/modalities\cite{cui2024novel,huang2024generalized,liang2024factorized,ren2024dynamic}.
Utilizing these unlabeled data while considering the need for data privacy among clients has given rise to an emerging field of federated multi-view clustering (FedMVC) \cite{chen2023federated}, which enables multiple clients to collaboratively train consistent clustering models without exposing private data. The clustering models across distributed clients can be applied in many applications (e.g., recommendation \cite{huang2020federated} and medicine \cite{che2022federated}) and thus FedMVC attracts increasing research interest \cite{hu2023efficient,lin2023federated,qiao2024federated,jiang2024heterogeneity}.

Existing FedMVC methods usually assume that clients are isomorphic and all of them belong to either single-view clients or multi-view clients.
For instance, FedDMVC \cite{chen2023federated} assumes that a dataset with $V$ views is distributed across $V$ single-view clients, each having the same set of samples. 
FedMVFPC \cite{hu2023efficient} assumes that the data are distributed among multiple multi-view clients, with each client having $V$ views and non-overlapping samples among clients.
Despite the achieved success, they still have limitations when handling some practical FedMVC scenarios with heterogeneous hybrid views, where different clients are heterogeneous such that single-view clients and multi-view clients are hybrid.

This scenario is prevalent in real world situations \cite{chen2022fedmsplit,yu2022multimodal},
for example, hospitals in metropolitan areas employ CT, X-ray, and EHR for disease detection, whereas remote areas usually rely on a single detection method.
Similarly, smartphones can simultaneously capture audio and images, but recording devices are limited to collecting audio data only.

The presence of heterogeneous hybrid views might limit the applicability of previous FedMVC methods. We decompose these challenges into two issues. 
(a) \emph{Client gap}. Multi-view clients collect multiple views that have the opportunity to learn comprehensive cluster partitions by leveraging multi-view information.
Conversely, single-view clients only own single views and easily obtain biased cluster partitions if the single view only contains marginal information.
(b) \emph{View gap}. Hybrid views collected by different clients have quality differences (e.g., images might contain richer visual information than texts, but texts could have semantic information), and extracting complementary information from different views across all clients is not trivial.

\begin{figure*}[!t]
    \centering
    \includegraphics[width=14cm]{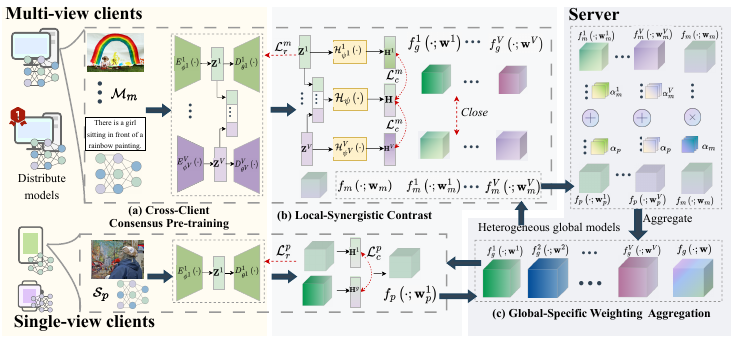}
    \caption{The framework of FMCSC. Initially, each client conducts cross-client consensus pre-training to alleviate model misalignment (Section 3.2). Then, all clients begin training using the designed local-synergistic contrast (Section 3.3) and upload their local models to the server. The server performs global-specific weighting aggregation and distributes multiple heterogeneous global models to all clients (Section 3.4). Finally, leveraging global models received from the server, clients discover complementary cluster structures across all clients.}
    \label{fig:framework}
\end{figure*}

In this paper, we introduce a novel FedMVC method, namely Federated Multi-view Clustering via Synergistic Contrast (FMCSC), which can simultaneously leverage the single-view and multi-view data across heterogeneous clients to mine clustering structures from hybrid views.
Figure \ref{fig:framework} shows an overview of our FMCSC framework.
First,
we note that lacking unified information supervision, naive aggregation of local models may lead to model misalignment.
Therefore, we propose cross-client consensus pre-training to align the local models on all clients to avoid their misalignment.
Second,
to address the client gap, we design local-synergistic contrastive learning that helps to mitigate the heterogeneity between single-view clients and multi-view clients.
In particular, we leverage feature-level and model-level contrastive learning to align multi-view clients and single-view clients, respectively.
Third, to tackle the view gap, we develop the global-specific weighting aggregation which encourages global models to learn robust features from hybrid views, further exploring complementary cluster structures.
The local-synergistic contrastive learning and global-specific weighting aggregation promote each other to explore the data cluster structures distributed on multiple clients.
Overall, FMCSC effectively facilitates all clients in bridging the client gap and view gap within the heterogeneous hybrid views through theoretical and experimental analysis.

Our main contributions are as follows:
 \begin{itemize}
     \item We propose a novel FedMVC method that can handle the heterogeneous hybrid views and explain the success mechanism of the proposed method through theoretical analysis from the perspective of bridging client and view gaps. 
    \item We design local-synergistic contrastive learning and global-specific weighting aggregation, using mutual information as a bridge to connect local and global models, together to explore the cluster structures in multi-view data distributed on different clients.
    \item Theoretical and experimental analyses verified the effectiveness of FMCSC, which shows excellent clustering performance under various federated learning scenarios.
 \end{itemize}

\section{Related Work}
Federated multi-view clustering (FedMVC) has emerged to explore cluster structures in multi-view data distributed on multiple clients. Existing FedMVC methods can be classified into two categories based on the partitioning of multi-view data among clients. 
(1) Vertical FedMVC assumes that a dataset with $V$ views is distributed across $V$ single-view clients, each having the same set of samples. Robust federated multi-view learning (FedMVL) \cite{huang2022efficient} addresses high communication costs, fault tolerance, and stragglers. Federated deep multi-view clustering (FedDMVC) \cite{chen2023federated} focuses on addressing the challenges of feature heterogeneity and incompleteness. 
Existing vertical FedMVC methods still rely on the idealistic assumption that different views of the same sample can be aligned across clients, which warrants further investigation.
(2) Horizontal FedMVC assumes that the data are distributed among multiple multi-view clients, with each client having $V$ views and non-overlapping samples among clients. Federated multi-view fuzzy c-means consensus prototypes clustering (FedMVFPC) \cite{hu2023efficient} utilizes federated learning mechanisms to perform fuzzy c-means clustering on multi-view data. Horizontal federated multi-view learning (H-FedMV) \cite{che2022federated} aims to improve the local disease prediction performance by sharing training models among clients. 
Although existing approaches have been successful, they still have limitations in handling practical FedMVC scenarios with heterogeneous hybrid views.
Specifically, heterogeneity refers to the differences among clients, where single-view and multi-view clients coexist. The hybrid views indicate the uncertainty in the number and quality of views involved in training.
Our proposed FMCSC is a variant of horizontal FedMVC, and can handle such scenarios by bridging the client gap and the view gap among clients.

\section{Methodology}

The key goal of FMCSC is to bridge client and view gaps. 
On the one hand, multi-view clients have the opportunity to learn comprehensive cluster partitions by leveraging multi-view information. Single-view clients aim to bridge the client gap between themselves and multi-view clients, thereby avoiding obtaining biased clustering partitions.
On the other hand, considering the inherent discrepancies in data quality among different views, our objective is to extract complementary information from hybrid views across all clients.

\subsection{Problem Definition}
We consider a heterogeneous federated learning setting where $M$ multi-view clients and $S$ single-view clients collaborate to train and mine complementary clustering structures using multiple heterogeneous global models $\left\{f_{g}^1\left(\cdot; \mathbf{w}^{1}\right),\ldots,f_{g}^V\left(\cdot; \mathbf{w}^{V}\right),f_{g}\left(\cdot; \mathbf{w}\right) \right\}$. 
Here, $f_{g}^v\left(\cdot; \mathbf{w}^{v}\right)$ represents the global model capable of handling the $v$-th view type, and $f_{g}\left(\cdot; \mathbf{w}\right)$ represents the global model capable of handling multi-view data.
Each single-view client $p \in[S]$ has its private dataset $\mathcal{S}_{p}=\left\{\mathbf{x}_{i}^{v}\right\}_{i=1}^{\left|\mathcal{S}_{p}\right|}$, where $\mathbf{x}_{i}^{v}$ represents the $i$-th sample of the $p$-th single-view client collected from the $v$-th view type. It adopts a small model $f_{p}\left(\cdot ; \mathbf{w}_{p}^v\right): \mathbb{R}^{D_v} \rightarrow \mathbb{R}^{d}$ based on its local view types. 
The multi-view client $m \in[M]$ has its local dataset $\mathcal{M}_{m}=\left\{\left(\mathbf{x}_{i}^{1}, \mathbf{x}_{i}^{2}, \ldots,\mathbf{x}_{i}^{V}\right)\right\}_{i=1}^{\left|\mathcal{M}_{m}\right|}$, where $\left(\mathbf{x}_{i}^{1}, \mathbf{x}_{i}^{2}, \ldots,\mathbf{x}_{i}^{V}\right)$ represents the $i$-th sample of the $m$-th multi-view client collected from $V$ different view types. It trains a small model $f_{m}\left(\cdot ; \mathbf{w}_{m}\right): \mathbb{R}^{\sum_{v=1}^{V}{D_v}}\rightarrow \mathbb{R}^{d}$ based on its local data. 
For simplicity, we assume that the output features of all models have the same dimension $d$.

\subsection{Cross-Client Consensus Pre-training}\label{sec:3.2}
Multi-view datasets often contain redundancy and random noise. Current mainstream methods employ self-supervised autoencoder models, such as autoencoder (AE) \cite{hinton1993autoencoders} and variational autoencoder (VAE) \cite{kingma2013auto} to learn high-level features from raw data. 
In FMCSC, we employ an encoder-decoder pair, denoted $E_{\phi ^v}^{v}(\cdot)$ and $D_{\theta ^v}^{v}(\cdot)$ with learnable parameters $\phi^{v}$ and $\theta^{v}$, for each view in each client. 
For multi-view client $m$, we define $\mathbf{z}_{i}^{v}=E_{\phi ^v}^{v}\left(\mathbf{x}_{i}^{v}\right) \in \mathbb{R}^{d_v}$ as the $d_v$-dimensional latent feature of the $i$-th sample, and the output of the autoencoder is $\hat{\mathbf{x}}_{i}^{v}=D_{\theta ^v}^{v}\left(\mathbf{z}_{i}^{v}\right) \in \mathbb{R}^{D_v}$. 
We calculate the reconstruction loss between the input $\mathbf{x}_{i}^{v}$ and the output $\hat{\mathbf{x}}_{i}^{v}$ for all samples in this client.
Additionally, the local model of this client consists of $V$ encoder-decoder pairs, which can be pre-trained by minimizing the reconstruction objective:
\begin{equation}\label{eq1}
    \mathcal{L}_{r}^{m}=\frac{1}{{\left|\mathcal{M}_{m}\right|}}\sum_{v=1}^{V}\sum_{i=1}^{\left|\mathcal{M}_{m}\right|}\left\|\mathbf{x}_{i}^{v}-D_{\theta ^v}^{v}\left(E_{\phi ^v}^{v}\left(\mathbf{x}_{i}^{v}\right)\right)\right\|^{2}_{2}.
\end{equation}

Similarly, for single-view client $p$ that contains a type of view data locally, it is sufficient to construct an encoder-decoder pair. Pre-training can be performed using the same objective as in Eq. (\ref{eq1}):
\begin{equation}\label{eq2}
    \mathcal{L}_{r}^{p}=\frac{1}{{\left|\mathcal{S}_{p}\right|}}\sum_{i=1}^{\left|\mathcal{S}_{p}\right|}\left\|\mathbf{x}_{i}^{v}-D_{\theta ^v}^{v}\left(E_{\phi ^v}^{v}\left(\mathbf{x}_{i}^{v}\right)\right)\right\|^{2}_{2}.
\end{equation}

\textbf{Key Observations:} In federated learning, diverse local data distributions frequently lead to model drift, causing slow and unstable convergence \cite{karimireddy2020scaffold,yu2022multimodal}. 
In FMCSC, single-view clients have never encountered data from other view types, thus intensifying the issue of model drift due to heterogeneous hybrid views. 
Moreover, the absence of uniformly labeled data across all clients allows the reconstruction objective of autoencoders to optimize from multiple different directions. 
In feature space, this problem manifests itself as angular deviations among features \cite{zhang2023federated}, and from the perspective of model aggregation, it manifests itself as model misalignment.
In other words, direct aggregation of models can blur the feature distinctions captured by local models, leading to inseparability among features, as shown in Figure \ref{visual}.

A direct strategy to alleviate model misalignment is through alignment. Based on this naive idea, we propose that the multi-view client that finishes training first should distribute its network parameters $\left\{E_{\phi ^1}^{1}(\cdot),\ldots, E_{\phi ^V}^{V}(\cdot) \right\}$ and $\left\{D_{\theta ^1}^{1}(\cdot),\ldots, D_{\theta ^V}^{V}(\cdot) \right\}$ to the remaining clients. 
Each client then performs pre-training based on this model, thereby alleviating the model misalignment caused by unsupervised training.
Notably, as pre-training solely involves training the autoencoder, the construction of local models is inherently dependent on the view type. 
Therefore, single-view clients can still refer to the network parameters of multi-view clients. This process facilitates consensus pre-training among the clients, ultimately leading to the uploading of pre-trained model parameters to the server. The server initializes global models based on the models uploaded by clients.

\subsection{Local-Synergistic Contrast}\label{sec:3.3}
During pre-training, the features extracted through the reconstruction objective usually contain both common semantics and view-private information. The latter is often meaningless or even misleading, leading to poor clustering effectiveness. 
To mitigate the adverse effects of view-private information, each client also needs to design a consistent objective during training to learn common semantics. 

For multi-view client $m$, it possesses information from multiple views. Inspired by \cite{wu2024self,yan2023gcfagg}, we employ feature contrastive learning to achieve consistency objectives. 
Considering the conflict between consistency and reconstruction objectives, we opt to operate in distinct feature spaces. 
We refer to the features obtained through the autoencoders $\left\{\left(\mathbf{z}_{i}^{1}, \mathbf{z}_{i}^{2}, \ldots,\mathbf{z}_{i}^{V}\right)\right\}_{i=1}^{\left|\mathcal{M}_{m}\right|}$ as low-level features. Moreover, we stack $V$ non-linear mappings $\left\{\mathcal{H}^1\left(\mathbf{Z}^{1}; \Psi^{1}\right),\ldots,\mathcal{H}^V\left(\mathbf{Z}^{V}; \Psi^{V}\right) \right\}$ to obtain high-level features $\left\{\left(\mathbf{h}_{i}^{1}, \mathbf{h}_{i}^{2}, \ldots,\mathbf{h}_{i}^{V}\right)\right\}_{i=1}^{\left|\mathcal{M}_{m}\right|}$. Additionally, a non-linear mapping $\mathcal{H}\left(\mathbf{Z}; \Psi\right) : \mathbb{R}^{\sum_{v=1}^{V}{d_v}}\rightarrow \mathbb{R}^{d}$ is constructed by:
\begin{equation}\label{eq3}
    \mathbf{H}=\mathcal{H}\left(\mathbf{Z}; \Psi\right) =\mathcal{H}\left(\left[\mathbf{Z}^{1}, \mathbf{Z}^{2}, \ldots, \mathbf{Z}^{V}\right]; \Psi\right),
\end{equation}
where $\left\{\mathbf{h}_{i}\right\}_{i=1}^{\left|\mathcal{M}_{m}\right|}=\mathbf{H} \in \mathbb{R}^{{\left|\mathcal{M}_{m}\right|} \times d}$, and $\mathbf{Z} \in \mathbb{R}^{{\left|\mathcal{M}_{m}\right|} \times {\sum_{v=1}^{V}{d_v}}}$. 
We aim to preserve the representative capacity of low-level features to prevent model collapse while learning the common semantics $\mathbf{H}$ among views in the high-level feature space.

In the high-level feature space, the common semantics $\mathbf{H}$ are learned from all views and should be very similar to the common semantics learned from individual views. 
Based on this, we define $\left\{\mathbf{h}_{i}, \mathbf{h}_{j}^{v}\right\}_{j=i}^{v=1, \ldots, V}$ as $V$ positive feature pairs, and the remaining $\left\{\mathbf{h}_{i}, \mathbf{h}_{j}^{v}\right\}_{j \ne i}^{v=1, \ldots, V}$ are $V({\left|\mathcal{M}_{m}\right|} -1)$ negative feature pairs. 
Then, we use cosine similarity to measure the similarity of feature pairs:
\begin{equation}\label{eq4}
    sim\left(\mathbf{h}_{i}, \mathbf{h}_{j}^{v}\right)=\frac{\left\langle\mathbf{h}_{i}, \mathbf{h}_{j}^{v}\right\rangle}{\left\|\mathbf{h}_{i}\right\|\left\|\mathbf{h}_{j}^{v}\right\|},
\end{equation}
where $\langle\cdot, \cdot\rangle$ is dot product operator. 
We introduce the temperature parameter $\tau_m$ to moderate the effect of similarity. Subsequently, the feature contrastive loss is formulated as:
\begin{equation}\label{eq5}
    \mathcal{L}_{c}^{m}=-\frac{1}{{\left|\mathcal{M}_{m}\right|}} \sum_{v=1}^{V}\sum_{i=1}^{{\left|\mathcal{M}_{m}\right|}} \log \frac{e^{sim\left(\mathbf{h}_{i}, \mathbf{h}_{i}^{v}\right) / \tau_m}}{\sum_{j\ne i}  e^{sim\left(\mathbf{h}_{i}, \mathbf{h}_{j}^{v}\right) / \tau_m}} .
\end{equation}

Moreover, multi-view clients aim to assist single-view clients in bridging client gaps and discarding view-private information detrimental to clustering. 
They optimize multiple heterogeneous global models using local data, ensuring that even the global model designed for single-view processing acquires generalized common semantics. 
Such common semantics are also advantageous for uncovering complementary clustering structures across clients. Concretely,  
\begin{equation} \label{eq6}
    \min_{\left\{ \mathbf{w}_m^v \right\} _{v=1}^{V}} {\sum_{v=1}^V{\left\| f_{m}^v\left(\cdot; \mathbf{w}_m^{v}\right)-f_{m}\left(\cdot; \mathbf{w}_{m}\right) \right\| _{2}^{2}}},
\end{equation}
where $f_{m}^v\left(\cdot; \mathbf{w}_m^{v}\right)$ represents the local model that can handle the $v$-th type of view, which is initialized by the global model $f_{g}^v\left(\cdot; \mathbf{w}^{v}\right)$. $f_{m}\left(\cdot; \mathbf{w}_{m}\right)$ represents the local model of multi-view client $m$, where
multi-view clients possess data of all view types. 

For single-view client $p$, where each sample only has a single view, we design a model contrastive learning to achieve consistency objectives. 
Specifically, client  $p$ contains data of the $v$-th view type $\left\{\mathbf{x}_{i}^{v}\right\}_{i=1}^{\left|\mathcal{S}_{p}\right|}$, with its local model as $f_{p}\left(\cdot; \mathbf{w}^{v}\right)$. 
To explore common semantics, we adopt the same approach as multi-view clients by constructing two non-linear mappings $\mathcal{H}^v\left(\mathbf{Z}^{v}; \Psi^{v}\right)$ and $\mathcal{H}\left(\mathbf{Z}; \Psi\right)$ to obtain high-level features $\mathbf{H}^{v}$ and common semantics $\mathbf{H}$. 
The global model$f_{g}\left(\cdot; \mathbf{w}^{v}\right)$ after aggregation further enhances its ability to learn generalized common semantics.
To encourage the local model of client $p$ to approach the more generalized global model, we formulate the model contrastive loss:
\begin{equation} \label{eq7}
    \mathcal{L}_{c}^{p} = -\frac{1}{{\left|\mathcal{S}_{p}\right|}} \sum_{i=1}^{{\left|\mathcal{S}_{p}\right|}} \log \frac{e^{sim\left(\mathbf{h}_{i}, \mathbf{h}_{i}^{g}\right) / \tau_p}}{e^{sim\left(\mathbf{h}_{i}, \mathbf{h}_{i}^{g}\right) / \tau_p}+e^{sim\left(\mathbf{h}_{i}, \mathbf{z}_{i}^{v}\right) / \tau_p}},
\end{equation}
where $\left\{\mathbf{h}_{i}^{g}\right\}_{i=1}^{\left|\mathcal{S}_{p}\right|}$ represents the output of the local data after being processed by the global model, and $\tau_p$ denotes the temperature parameter. The significance of Eq. (\ref{eq7}) lies in treating $\left\{\mathbf{h}_{i},\mathbf{h}_{i}^g\right\}_{i=1}^{\left|\mathcal{S}_{p}\right|}$ as positive pairs and $\left\{\mathbf{h}_{i},\mathbf{z}_{i}^v\right\}_{i=1}^{\left|\mathcal{S}_{p}\right|}$ as negative pairs. 
This allows the local model of client $p$ to converge towards the global model while amplifying the differences between the reconstruction and consistency objectives in the local model.

During training, the respective total losses for multi-view client $m$ and single-view client $p$ are:
\begin{equation} \label{eq8}
    \mathcal{L} ^m=\mathcal{L} _{r}^{m}+ \mathcal{L} _{c}^{m}, ~~~~~   \mathcal{L} ^p=\mathcal{L} _{r}^{p}+ \mathcal{L} _{c}^{p}.
\end{equation}
In the optimization of FMCSC, $\mathcal{L} _{r}^{m}$ and $\mathcal{L} _{r}^{p}$ are utilized as reconstruction losses to learn representations for each view individually. 
Meanwhile, $\mathcal{L} _{c}^{m}$ and $\mathcal{L} _{c}^{p}$ are employed to discover common semantics across views, facilitating the exploration of complementary clustering structures across clients.

\subsection{Global-Specific Weighting Aggregation}\label{sec:3.4}
To bridge the view gap and aggregate heterogeneous models, we design a weighted specific aggregation strategy on the server, yielding multiple heterogeneous global models.
\begin{theorem}\label{th1}
Assume $\delta_m, \delta_p\in(0,1)$ such that $p\left(\mathbf{h}_{i}^v \mid \mathbf{h}_{i} \right)>\delta_m$, $i=1,2, \cdots, \left|\mathcal{M}_{m}\right| $ and $p\left(\mathbf{h}_{i}^g \mid \mathbf{h}_{i} \right)=p\left(\mathbf{z}_{i}^v \mid \mathbf{h}_{i} \right)>\delta_p$, $i=1,2, \cdots, \left|\mathcal{S}_{p}\right| $ hold. The following inequality establishes the relationship between the consistency objectives and the mutual information of the multi-view client $m$ and single-view client $p$, respectively:
\begin{equation} \label{eq9}
    \begin{aligned}
    &\sum_{v=1}^{V}I\left(\mathbf{H}, \mathbf{H}^{v}\right) \geq V \log \left|\mathcal{M}_{m}\right|-\delta_m \mathcal{L}_{c}^{m},\\
    &I\left(\mathbf{H}, \mathbf{H}^{g}\right)-I\left(\mathbf{H}, \mathbf{Z}^{v}\right) \leq -2\delta_p \mathcal{L}_{c}^{p}.
    \end{aligned}
\end{equation}
\end{theorem}
Proofs of all the theorems in this paper are provided in Appendix \ref{proof} due to space limit. Theorem \ref{th1}, Eq. (\ref{eq5}) and Eq. (\ref{eq7}) indicate that minimizing contrastive loss $\mathcal{L}_{c}^{m}$ and $\mathcal{L}_{c}^{p}$ are equal to maximizing mutual information. Such connection has also been discussed in \cite{xu2023self,zhong2020deep}.

Through theoretical analysis, we use mutual information $ \sum_{v=1}^{V}I\left(\mathbf{H}, \mathbf{H}^{v}\right)$ and $I\left(\mathbf{H}, \mathbf{H}^{g}\right)-I\left(\mathbf{H}, \mathbf{Z}^{v}\right)$ as weights to evaluate the quality of the models from multi-view clients and single-view clients, respectively. 
A higher level of mutual information indicates better model quality, leading to higher weights during aggregation.

Considering the heterogeneity of the models, we aggregate client models with the same architecture on the server, referred to as specific aggregation. Specifically,
\begin{equation} \label{eq10}
    \begin{aligned}
    &\ f_{g}\left(\cdot; \mathbf{w}\right) =\sum_{m=1}^{M}\alpha_m f_{m}\left(\cdot; \mathbf{w}_m\right),\\
    &\ f_{g}^v\left(\cdot; \mathbf{w}^{v}\right)=\sum_{m=1}^{M}\alpha_m^v f_{m}^v\left(\cdot; \mathbf{w}_m^{v}\right)+\sum_{p=1}^{S}\alpha_p f_{p}\left(\cdot; \mathbf{w}_p^{v}\right), \\
    \end{aligned}
\end{equation}
where $v=1,2, \cdots, V$, $\alpha_m$ and $\alpha_p$ represent the weights for model aggregation of multi-view client $m$ and single-view client $p$ respectively.
In this scenario, there are a total of $M$ multi-view clients and $S$ single-view clients, resulting in $(V+1)$ heterogeneous global models.

For global models that handle multiple view types simultaneously, the expected risk is defined as $\mathcal{L} _M(f)$ and optimized by minimizing the empirical risk $\widehat{\mathcal{L} }_M(f)$. Similarly, for global models dealing with processing a single view type, such as the $v$-th view type, the expected and empirical risks are defined as $\mathcal{L}_{S^v}(f)$ and $\widehat{\mathcal{L} }_{S^v}(f)$ respectively. Inspired by previous works \cite{li2021sharper,tang2022deep,xu2022self} on the generalization bound of clustering approaches, we obtain the following theorem by analyzing the generalization bound of the proposed FMCSC method.
\begin{theorem} \label{th2}
    Suppose that for any $\mathbf{x} \in \mathcal{X}$ and $f \in \mathcal{F}$, there exists $D<\infty$ such that $\left\|\mathbf{x}\right\|,\left\|f_x\left(\mathbf{x}\right)\right\|,\left\|f_{z}^v\left(\mathbf{x}\right)\right\|,\left\|f_{h}^v\left(\mathbf{x}\right)\right\|,\left\|f_{h}^0\left(\mathbf{x}\right)\right\| \in[0, D]$ hold.
    With probability $1-\delta$ for any $f \in \mathcal{F}$, the following inequality holds
    \begin{equation} \label{eq11}
    \begin{aligned}
    \mathcal{L}_M(f) \leq  & \widehat{\mathcal{L}}_M(f)+\frac{12V D^{2}}{\sqrt{M\left| \mathcal{M} _m \right|}}+9V D^{2} \sqrt{\frac{\log \frac{1}{\delta}}{2 M\left| \mathcal{M} _m \right|}}, \\
    L_{S^v}(f)\leq & \widehat{L}_{S^v}(f) + \frac{10 D^{2}}{\sqrt{S^v\left| \mathcal{S} _p \right|}}+8 D^{2} \sqrt{\frac{\log \frac{1}{\delta}}{2 S^v\left| \mathcal{S} _p \right|}} \\
    & + \sqrt{\frac{4}{M\left| \mathcal{M} _m \right|}\left(d \log \frac{2 e M\left| \mathcal{M} _m \right|}{d}+\log \frac{4}{\delta}\right)}+d_{\mathcal{F}}\left(\tilde{\mathcal{D}}_{v}, \tilde{\mathcal{D}}\right)+\lambda_{v}, \\
    \end{aligned}
\end{equation}
where $M$ is the number of multi-view participating clients, $S^v$ is the number of single-view clients with $v$-th view type that participated in the training, $\left| \mathcal{M} _m \right|$ and $\left| \mathcal{S} _p \right|$ are the number of samples in each multi-view client and single-view client, respectively. $d_{\mathcal{F}}\left(\tilde{\mathcal{D}}_{v}, \tilde{\mathcal{D}}\right)$ measures the difference between data from the $v$-th view distribution $\tilde{\mathcal{D}}_{v}$ and the multi-view data distribution $\tilde{\mathcal{D}}$.
\end{theorem}

Three key implications can be derived from Theorem \ref{th2}: 
i) For global models capable of handling multi-view data, having more samples from multi-view clients, such as increasing the number of samples per client or adding multi-view participating clients, contributes to the improvement of generalization performance.
ii) For global models dealing with single-view data, having more samples from both multi-view and single-view clients enhances their generalization performance, which is a reflection of multi-view clients helping single-view clients to bridge the client gap.
iii) Although the view gap is mitigated, the high dissimilarity among views still leads to high distribution divergence $d_{\mathcal{F}}\left(\tilde{\mathcal{D}}_{v}, \tilde{\mathcal{D}}\right)$, which impairs the quality of the global model.

Finally, each client applies the $K$-means \cite{macqueen1967classification} on the common semantics $\mathbf{H}$ obtained from the corresponding global model to obtain their local clustering results. 
By concatenating clustering results from all clients, we obtain the overall clustering results. We define $\mathbf{U}$ as the cluster indicator matrix and $\mathbf{C}$ as the center matrix of clustering. Specifically, the common semantics $\mathbf{H}$ from the clients can be factorized as follows:
\begin{equation} \label{eq12}
    \begin{aligned}
&\min _{\mathbf{U}, \mathbf{C}}\|\mathbf{H}-\mathbf{U C}\|^{2} \\
& \text { s.t. } \mathbf{U} \mathbf{I}=\mathbf{I}, \mathbf{U} \geq \mathbf{0}.\\
\end{aligned}
\end{equation}

\section{Experiments}
\subsection{Experimental Settings}
\textbf{Datasets.} Our experiments are carried out on four multi-view datasets. Specifically, \textbf{MNIST-USPS} \cite{peng2019comic} comprises 5000 samples collected from two handwritten digital image datasets, which are considered as two views. 
\textbf{BDGP} \cite{cai2012joint} consists of 2500 samples across 5 drosophila categories, with each sample having textual and visual views. 
\textbf{Multi-Fashion} \cite{xiao2017fashion} contains images from 10 categories, where we treat three different styles of one object as three views, resulting in 10000 samples. 
\textbf{NUSWIDE} \cite{chua2009nus} consists of 5000 samples obtained from web images with 5 views. 
Considering the sample quantities, we allocate BDGP to 12 clients, MNIST-USPS and NUSWIDE to 24 clients respectively, and Multi-Fashion to 48 clients to simulate the federated learning settings.


\textbf{Comparison Methods.} We select 9 state-of-the-art methods, including HCP-IMSC \cite{li2022high}, IMVC-CBG \cite{wang2022highly}, DSIMVC \cite{tang2022deep}, LSIMVC \cite{liu2022localized}, ProImp \cite{li2023incomplete}, JPLTD \cite{lv2023joint}, CPSPAN \cite{jin2023deep}, FedDMVC \cite{chen2023federated} and FCUIF \cite{ren2024novel}. 
Among them, apart from FedDMVC and FCUIF, which are FedMVC methods, all the other comparison methods are centralized incomplete multi-view clustering methods. 
To ensure fair comparisons, we concatenate the data distributed among the clients and use them as the input for centralized methods.
Among these, the data from multi-view clients can be regarded as complete data, while the data from single-view clients can be considered as missing data.

\textbf{Implementation Details.} For an encoder-decoder pair, the encoder structure is $\text { Input- } \mathrm{Fc}_{500}-\mathrm{Fc}_{500}-\mathrm{Fc}_{2000}-\mathrm{Fc}_{20}$, and the decoder is symmetric with the encoder.
Also, we set temperature parameters $\tau_m=\tau_p=0.5$ and use the batch size of 256.
The output dimension $d$ is set to 20 for all local and global models and communication rounds $R$ is set to 5.
All experiments in the paper involving FMCSC that are not mentioned are performed when $M/S$ = 1:1.


\begin{table*}[!t]
 \caption{ Clustering results (mean±std\%) of all methods on four datasets. The best and second best results are denoted in \textbf{bold} and \underline{underline}.}
\centering
\begin{adjustbox}{width=1\textwidth}
\begin{tabular}{|c|r|c c c|c c c|c c c|}
\hline
\multicolumn{1}{|c}{} &
\multicolumn{1}{|r}{Multi- / Single- clients} & \multicolumn{3}{|c}{$M/S$ =  2:1} & \multicolumn{3}{|c}{$M/S$ =  1:1} & \multicolumn{3}{|c|}{$M/S$ =  1:2}           \\
\hline
\multicolumn{1}{|c}{} &
\multicolumn{1}{|r|}{Evaluation metrics} & ACC & NMI & ARI & ACC & NMI & ARI  & ACC & NMI & ARI              \\
\hline

\multirow{10}{*}{\rotatebox{90}{MNIST-USPS}} 
& HCP-IMSC (2022) \cite{li2022high} &80.2±0.0	&74.8±0.0	&69.8±0.1	&79.0±0.2	&71.6±0.2	&66.9±0.2	&76.2±0.1	&71.0±0.1	&63.6±0.1 \\
& IMVC-CBG (2022) \cite{wang2022highly} &46.6±0.1	&40.3±0.2	&22.2±0.4	&38.9±0.1	&35.3±0.3	&14.9±0.2	&37.3±0.6	&31.7±0.4	&10.6±0.2 \\
& DSIMVC (2022) \cite{tang2022deep} &55.1±0.1	&27.6±0.1	&25.0±0.1	&54.5±0.1	&26.7±0.1	&24.4±0.2	&54.1±0.2	&26.6±0.2	&24.0±0.3 \\
& LSIMVC (2022) \cite{liu2022localized} &59.3±0.2	&55.2±0.9	&38.2±1.7	&52.7±0.4	&46.5±0.2	&25.8±0.5	&41.8±0.4	&37.1±0.4	&14.2±0.5 \\
& ProImp (2023) \cite{li2023incomplete} &\underline{91.2±0.9}	&\underline{84.4±1.0}	&\underline{80.8±2.1}	&\underline{87.3±0.6}	&\underline{77.9±0.7}	&\underline{73.1±1.0}	&\underline{84.8±0.9}	&\underline{75.9±0.9}	&\underline{66.6±1.8} \\
& JPLTD (2023) \cite{lv2023joint} &40.7±0.1	&22.6±0.1	&17.3±0.0	&40.0±0.2	&19.8±0.1	&15.2±0.1	&32.3±0.1	&13.7±0.1	&10.1±0.1 \\
& CPSPAN (2023) \cite{jin2023deep} &79.5±2.5	&77.3±2.0	&70.7±2.2	&76.5±2.7	&73.7±2.1	&66.6±3.0	&74.6±3.6	&74.5±3.3	&64.5±3.5 \\
& FedDMVC (2023) \cite{chen2023federated} &81.1±0.9	&81.9±0.9	&73.7±1.4	&69.9±1.5	&72.5±2.9	&58.9±2.6	&63.1±0.4	&62.6±0.5	&48.4±0.3 \\  
& FCUIF (2024) \cite{ren2024novel} &85.3±0.3	&83.2±0.3	&75.7±0.5	&72.8±1.2	&70.3±1.5	&64.4±1.8	&67.2±0.8	&64.4±0.9	&53.5±0.8 \\ 
& FMCSC (Ours) &\textbf{95.1±0.8}	&\textbf{87.8±1.2}	&\textbf{88.8±1.5}	&\textbf{92.9±1.2}	&\textbf{84.2±2.2}	&\textbf{85.0±2.5}	&\textbf{90.1±1.2}	&\textbf{79.4±2.6}	&\textbf{79.5±3.2} \\
\hline

\multirow{10}{*}{\rotatebox{90}{BDGP}} 
& HCP-IMSC (2022) \cite{li2022high} &93.1±0.0	&81.9±0.0	&83.6±0.0	&89.8±0.0	&73.4±0.1	&76.4±0.0	&\underline{89.5±0.0}	&\underline{72.5±0.1}	&\underline{76.7±0.0} \\
& IMVC-CBG (2022) \cite{wang2022highly} &37.9±0.4	&21.0±1.0	&10.4±0.1	&37.2±0.1	&21.0±0.0	&7.8±0.0	&36.9±0.1	&20.4±0.1	&6.4±0.0 \\
& DSIMVC (2022) \cite{tang2022deep} &92.5±0.4	&81.7±0.8	&84.9±0.9	&89.5±2.0	&76.5±2.0	&77.8±2.2	&86.1±3.3	&70.0±3.9	&76.6±3.4 \\
& LSIMVC (2022) \cite{liu2022localized} &44.1±0.5	&23.7±0.4	&5.7±0.2	&39.2±1.3	&19.7±0.5	&4.8±0.2	&35.3±1.6	&14.9±1.3	&2.8±0.4 \\
& ProImp (2023) \cite{li2023incomplete} &91.6±0.3 &\underline{82.4±3.8}		&80.0±0.9 &90.4±1.5	&76.2±0.5 &79.3±1.6	&75.6±0.5	&52.3±2.0	&44.6±1.8 \\
& JPLTD (2023) \cite{lv2023joint} &56.5±0.2	&41.3±0.1	&31.7±0.0	&49.4±0.1	&33.3±0.0	&18.5±0.0	&51.0±0.2	&34.1±0.1	&21.5±0.0 \\
& CPSPAN (2023) \cite{jin2023deep} &78.7±0.6	&58.3±1.3	&58.6±1.6	&57.3±1.3	&50.3±2.3	&39.4±3.7	&52.4±1.5	&34.7±1.4	&27.1±2.1 \\
& FedDMVC (2023) \cite{chen2023federated} &92.0±0.1	&80.2±0.2	&84.7±0.1	&\underline{91.5±0.5}	&\underline{77.1±0.4}	&\underline{80.3±0.7}	&82.2±0.2	&63.4±0.3	&61.9±0.3
 \\  
& FCUIF (2024) \cite{ren2024novel} &\underline{93.8±0.1}	&82.2±0.1	&\underline{85.1±0.1}	&90.3±0.2	&75.2±0.3	&78.4±0.3	&85.7±0.2	&67.5±0.3	&63.2±0.2
 \\ 
& FMCSC (Ours) &\textbf{94.5±0.8}	&\textbf{83.9±1.2}	&\textbf{86.8±1.5}	&\textbf{91.9±1.2}	&\textbf{77.3±2.2}	&\textbf{81.0±2.5}	&\textbf{90.0±1.2}	&\textbf{73.3±2.6}	&\textbf{76.8±3.2} \\
\hline

\multirow{10}{*}{\rotatebox{90}{Multi-Fashion}} 
& HCP-IMSC (2022) \cite{li2022high} &70.6±0.1	&67.4±0.1	&57.7±0.1	&67.1±0.1	&64.7±0.1	&53.1±0.1	&59.9±0.7	&56.4±0.9	&41.8±1.1 \\
& IMVC-CBG (2022) \cite{wang2022highly} &46.3±0.0	&49.4±0.0	&26.3±0.0	&43.2±0.1	&42.7±0.1	&19.2±0.1	&38.9±0.2	&39.4±0.3	&13.5±0.4 \\
& DSIMVC (2022) \cite{tang2022deep} &\underline{82.7±1.3}	&\underline{83.6±1.1}	&\underline{74.5±1.1}	&\underline{77.7±1.4}	&\underline{76.7±0.8}	&\underline{66.8±0.8}	&\underline{76.7±1.7}	&\underline{75.8±1.5}	&\underline{66.4±1.4}
 \\
& LSIMVC (2022) \cite{liu2022localized} &51.1±0.5	&49.9±0.1	&31.5±0.5	&50.2±0.6	&52.2±0.1	&35.2±0.1	&49.9±0.2	&48.6±0.0	&28.2±0.0
 \\
& ProImp (2023) \cite{li2023incomplete} &69.1±0.4	&66.3±0.3	&55.2±0.8	&69.0±0.1	&64.6±0.2	&52.5±0.2	&53.9±2.4	&50.7±1.5	&27.8±1.6 \\
& JPLTD (2023) \cite{lv2023joint} &44.6±0.0	&43.4±0.0	&36.9±0.1	&37.2±0.1	&36.6±0.1	&29.5±0.1	&25.8±0.1	&25.1±0.1	&16.5±0.1 \\
& CPSPAN (2023) \cite{jin2023deep} &64.1±1.2	&71.4±1.3	&55.8±1.5	&61.6±2.0	&69.7±1.3	&54.4±1.8	&59.3±2.0	&68.0±1.8	&53.3±1.9 \\
& FedDMVC (2023) \cite{chen2023federated} &67.7±0.3	&74.6±0.8	&58.0±0.6	&66.6±0.4	&65.3±0.7	&54.3±0.7	&57.6±0.7	&58.5±0.8	&43.2±0.7 \\ 
& FCUIF (2024) \cite{ren2024novel} &70.7±0.5	&79.4±0.5	&63.1±0.4	&68.4±0.6	&71.5±0.4	&59.2±0.5	&62.5±0.6	&61.3±0.5	&45.6±0.5 \\ 
& FMCSC (Ours) &\textbf{92.4±0.1}	&\textbf{85.8±0.2}	&\textbf{84.7±0.3}	&\textbf{90.4±0.6}	&\textbf{82.8±0.7}	&\textbf{80.9±1.0}	&\textbf{87.5±0.6}	&\textbf{79.1±0.1}	&\textbf{76.3±1.0}
 \\
\hline

\multirow{10}{*}{\rotatebox{90}{NUSWIDE}} 
& HCP-IMSC (2022) \cite{li2022high} &36.1±0.0	&8.5±0.0	&6.7±0.0	&35.3±0.1	&8.2±0.0	&6.3±0.0	&31.3±0.0	&6.0±0.1	&4.7±0.1 \\
& IMVC-CBG (2022) \cite{wang2022highly} &30.8±0.1	&4.8±0.0	&3.1±0.0	&30.4±0.0	&4.6±0.0	&2.5±0.0	&29.3±0.0	&4.0±0.0	&1.9±0.1 \\
& DSIMVC (2022) \cite{tang2022deep} &51.1±1.3	&25.3±0.8	&23.4±0.8	&50.6±0.8	&22.2±0.7	&20.4±0.6	&46.7±0.5	&18.3±0.6	&\underline{16.0±0.4} \\
& LSIMVC (2022) \cite{liu2022localized} &37.2±0.2	&10.8±0.1	&6.8±0.1	&36.4±0.3	&11.8±0.1	&7.0±0.4	&33.9±0.4	&9.2±0.3	&5.8±0.2 \\
& ProImp (2023) \cite{li2023incomplete} &38.4±0.1	&11.1±0.0	&8.3±0.1	&37.1±0.4	&10.5±0.1	&7.6±0.2	&34.3±0.7	&8.0±0.0	&6.1±0.0 \\
& JPLTD (2023) \cite{lv2023joint} &\underline{53.0±0.2}	&\underline{25.9±0.2}	&\underline{23.7±0.1}	&\underline{51.5±0.5}	&\underline{22.5±1.0}	&\underline{21.5±0.8}	&\underline{50.0±0.2}	&\underline{19.4±0.1}	&14.1±0.1 \\
& CPSPAN (2023) \cite{jin2023deep} &33.7±0.2	&9.0±0.9	&6.2±0.1	&33.3±0.3	&6.6±0.9	&4.4±0.1	&29.4±0.6	&5.4±1.1	&3.2±0.4 \\
& FedDMVC (2023) \cite{chen2023federated} &41.7±0.2	&14.4±0.1	&12.3±0.1	&37.5±0.7	&9.8±0.9	&7.8±0.5	&32.6±0.2	&5.8±0.2	&4.3±0.2 \\
& FCUIF (2024) \cite{ren2024novel} &45.2±0.3	&15.0±0.3	&14.1±0.2	&40.2±0.5	&10.0±0.6	&9.2±0.5	&38.2±0.4	&9.6±0.3	&8.2±0.3 \\  
& FMCSC (Ours) &\textbf{56.1±0.2}	&\textbf{26.3±0.5}	&\textbf{23.9±0.4}	&\textbf{52.7±0.2}	&\textbf{23.0±0.3}	&\textbf{21.8±0.4}	&\textbf{50.8±0.9}	&\textbf{20.1±0.7}	&\textbf{18.8±0.8}
 \\
\hline

\end{tabular}
\end{adjustbox}
 \label{result}
\end{table*}

\begin{table*}[!h]
\caption{Ablation studies on four datasets when $M/S$=1:1.}
\centering{
\resizebox{1\textwidth}{!}
{
\begin{tabular}{|c|c c c c|c c c|c c c|c c c|c c c| }
\hline
   \multicolumn{1}{|c|}{} & \multicolumn{4}{c|} {Components}  &\multicolumn{3}{c|}{MNIST-USPS}& \multicolumn{3}{c|}{BDGP}  & \multicolumn{3}{c|}{Multi-Fashion} & \multicolumn{3}{c|}{NUSWIDE}   \\
\hline
 \multicolumn{1}{|c|}{} & A  & B & C &D & ACC & NMI & ARI   & ACC & NMI  &ARI   &  ACC & NMI & ARI &  ACC & NMI & ARI \\
\hline
 \multicolumn{1}{|c|}{Item-1} &   & $\checkmark$ & $\checkmark$ & $\checkmark$ & 91.12	&82.24	&81.63	&69.84 	&48.71 	&46.29 	&87.80 	&79.86 	&76.61 	&42.16 	&12.37 	&11.16 \\

 \multicolumn{1}{|c|}{Item-2} & $\checkmark$  &  & $\checkmark$ & $\checkmark$ & 60.82 	&39.74 	&34.28 	&61.04 	&30.76 	&30.14 	&56.04 	&34.22 &27.75 	&42.08 	&10.15 	&9.06   \\

 \multicolumn{1}{|c|}{Item-3} & $\checkmark$  & $\checkmark$ &  & $\checkmark$ &88.66 	&76.85 	&76.84 	&87.40 	&68.96 	&71.09 	&82.58 	&74.29 	&68.82 	&45.86 	&15.06 	&12.64  \\

\multicolumn{1}{|c|}{Item-4} & $\checkmark$  & $\checkmark$ & $\checkmark$  &   & 89.52 	&78.12 	&78.31 	&85.44 	&64.15 	&67.26 	&88.14 	&79.61 	&76.91 	&47.54 	&15.20 	&13.79  \\

 \multicolumn{1}{|c|}{Item-5} & $\checkmark$  & $\checkmark$ & $\checkmark$ & $\checkmark$ & \textbf{93.38} 	&\textbf{84.98} 	&\textbf{85.96} 	&\textbf{91.92} 	&\textbf{77.29} 	&\textbf{80.95} 	&\textbf{90.36} 	&\textbf{82.81} 	&\textbf{80.89} 	&\textbf{54.04}	&\textbf{22.97}	&\textbf{21.81} \\
\hline

\end{tabular}
}
}
 \label{tab:ablation}
\end{table*}

\subsection{Results and Analysis}

\textbf{Clustering Results.} Table \ref{result} presents a quantitative comparison under various heterogeneous hybrid view scenarios. Each experiment is independently conducted five times, reporting average values and standard deviations. We construct different scenarios by adjusting the ratio of multi-view clients to single-view clients.
Remarkably, FMCSC achieves exceptional performance in various heterogeneous hybrid view scenarios, surpassing recent methods. 
This indicates our ability to achieve satisfactory clustering performance while preserving data privacy.
Furthermore, with the increasing proportion of single-view clients, all methods exhibit varying degrees of performance decline, aligning with common expectations. 
Even when the number of single-view clients is twice that of multi-view clients, FMCSC still achieves superior clustering performance, which is encouraging.

\textbf{Ablation Studies.} 
Components A and D represent the consensus pre-training process and weighted aggregation, respectively. Component B indicates that multi-view clients bring the global models closer as Eq. (\ref{eq6}). Component C indicates that model comparison in single-view clients as Eq. (\ref{eq7}). 
Table \ref{tab:ablation} shows that the impact of component A on clustering performance is dependent on the dataset. 
Additionally, note that the results in Item-1 are obtained after running four times the number of communication rounds compared to FMCSC. Although the initial misalignment of the model on MNIST-USPS and Multi-Fashion can be mitigated through multiple communication rounds. 
Component A still plays a crucial role in our training process, facilitating consensus among clients 
\begin{wrapfigure}{r}{0.3\textwidth}
  \centering
{\includegraphics[width=0.3\textwidth]{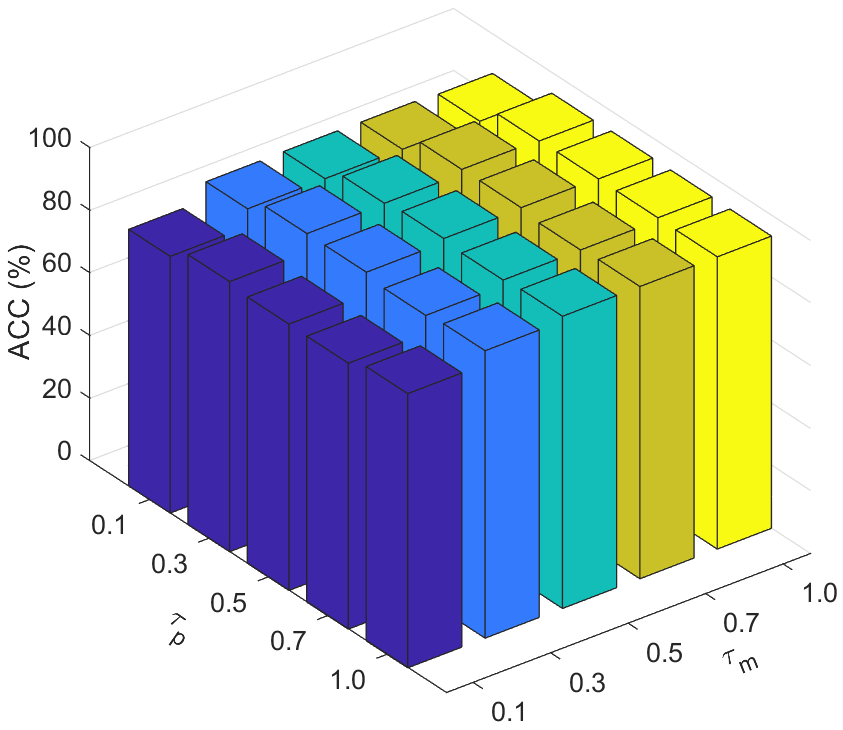}}
  \caption{ACC vs. $\tau_m$ and $\tau_p$.}
  \label{para}
\end{wrapfigure}
during pre-training to effectively alleviate model misalignment and accelerate convergence.
Item-3, 
which lacks component C, involves both single-view and multi-view clients carrying out the same 
operation of replacing their local models with global models. 
We observe a substantial improvement in Item-3 compared to Item-2, indicating that the success of model comparison in component C
is attributed to high-quality global models. 
This achievement requires collaborative efforts from both multi-view and single-view clients. 
Additionally, the inclusion of the weighted aggregation in component D enhances the benefits of collaborative training among all clients.

\textbf{Parameter Analysis.} We investigate the sensitivity of our clustering performance on MNIST-USPS dataset to two primary 
hyperparameters in local-synergistic contrastive learning: $\tau_m$ and $\tau_p$, as shown in Figure \ref{para}. The values of $\tau_m$ and $\tau_p$ are tuned within the range of {0.1, 0.3, 0.5, 0.7, 1}.
Our observations include: i) When $\tau_m$ and $\tau_p$ are set to small values, such as 0.1, the clustering performance of the proposed FMCSC decreases. This may be attributed to an excessive emphasis on view consistency, potentially resulting in an inseparable intrinsic feature space. 
ii) As the values of $\tau_m$ and $\tau_p$ increase, the clustering results gradually recover, and they exhibit insensitivity within the range of 0.3 to 1. Empirically, we set $\tau_m$ = $\tau_p$ = 0.5 for all datasets.

\textbf{Qualitative Study on Model Misalignment.}
To further quantify the impact of model misalignment and the effectiveness of our proposed strategy, we visualize the model outputs, i.e., the consensus semantics $\mathbf{H}$, both without consensus pre-training and with consensus pre-training. 
\begin{wrapfigure}{r}{0.51\textwidth}
  \centering
 \subfigure[Without consensus.]{\includegraphics[width=0.23\textwidth]{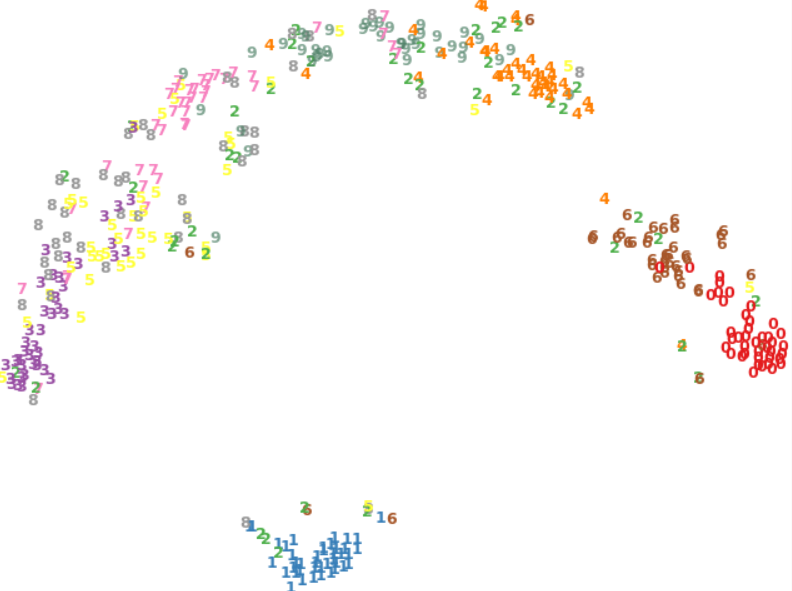}}
  \subfigure[With consensus.]{\includegraphics[width=0.23\textwidth]{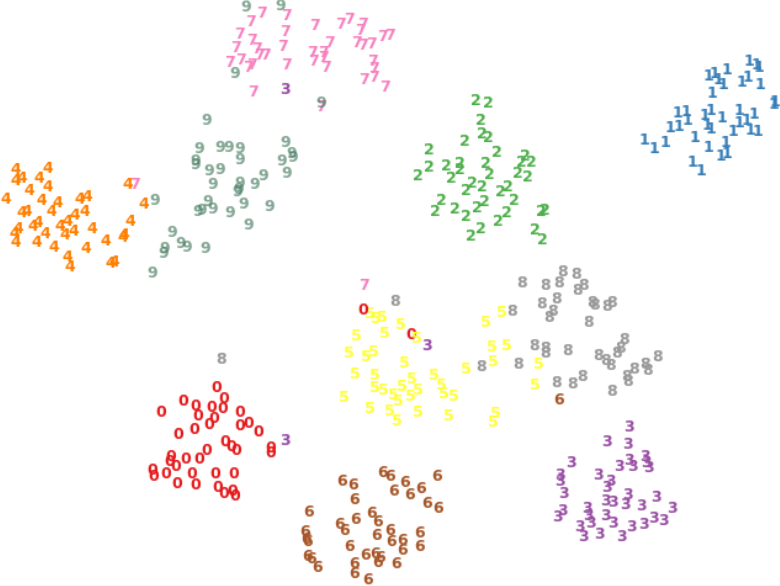}}
  \caption{Visualization on model misalignment.}
  \label{visual}
\end{wrapfigure}
Figure \ref{visual} displays the t-SNE \cite{van2008visualizing} visualizations generated from a randomly selected multi-view client, where different colors represent different classes. 
In Figure \ref{visual} (a), we observe that the output of the global model without consensus pre-training shows mixed features that cannot be clearly distinguished.
This confirms our viewpoint that direct parameter aggregation leads to model misalignment, manifested as feature confusion in the feature space. 
In contrast, Figure \ref{visual} (b) demonstrates that FMCSC produces more distinct and separable features, mitigating the negative impact of model aggregation.

\begin{figure*}[!h]
  \centering
  \subfigure[ACC $vs.$ Samples per client.]{\includegraphics[width=0.32\textwidth]{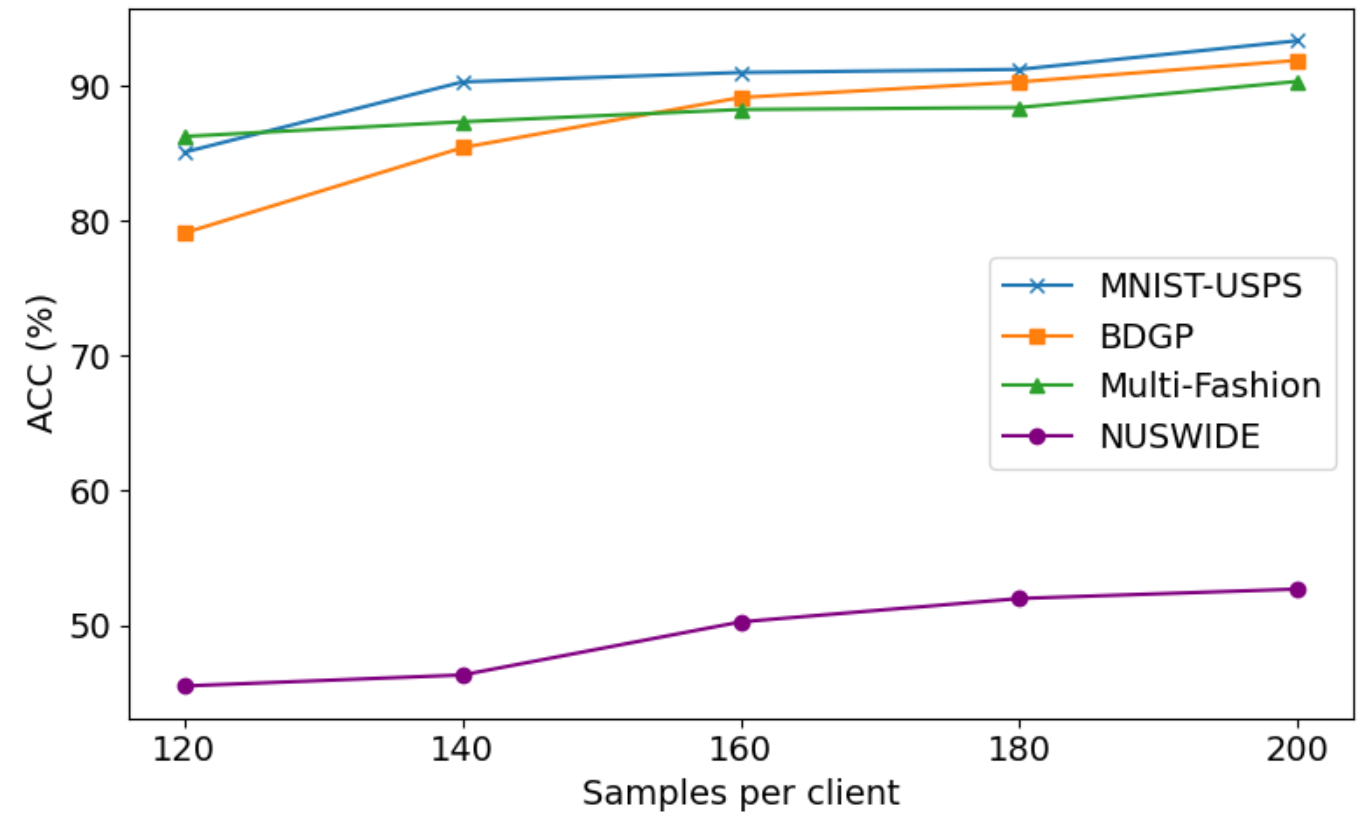}}
  \subfigure[ACC $vs.$ Number of clients.]{\includegraphics[width=0.32\textwidth]{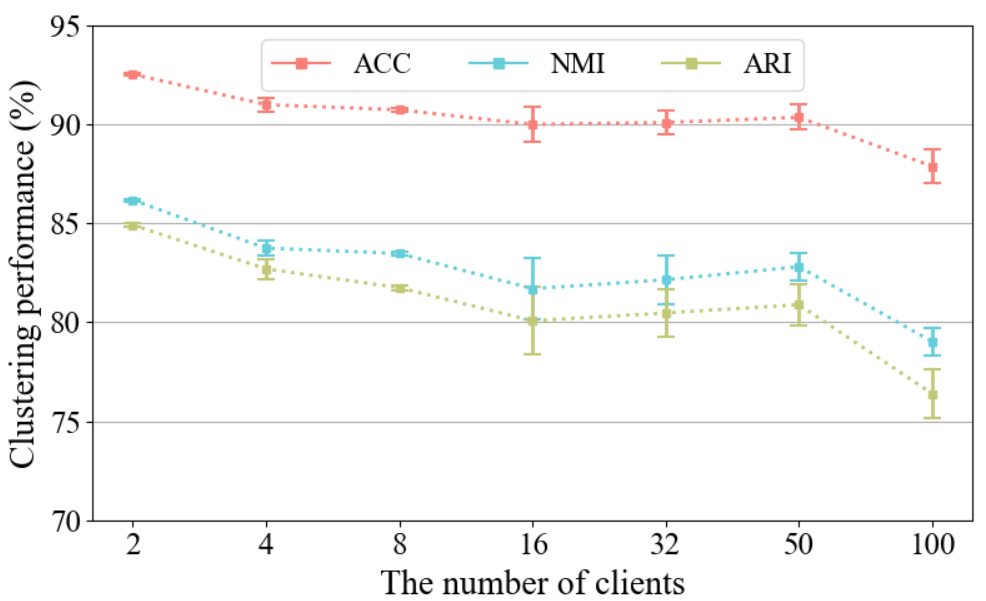}}
  \subfigure[Privacy analysis.]{\includegraphics[width=0.32\textwidth]{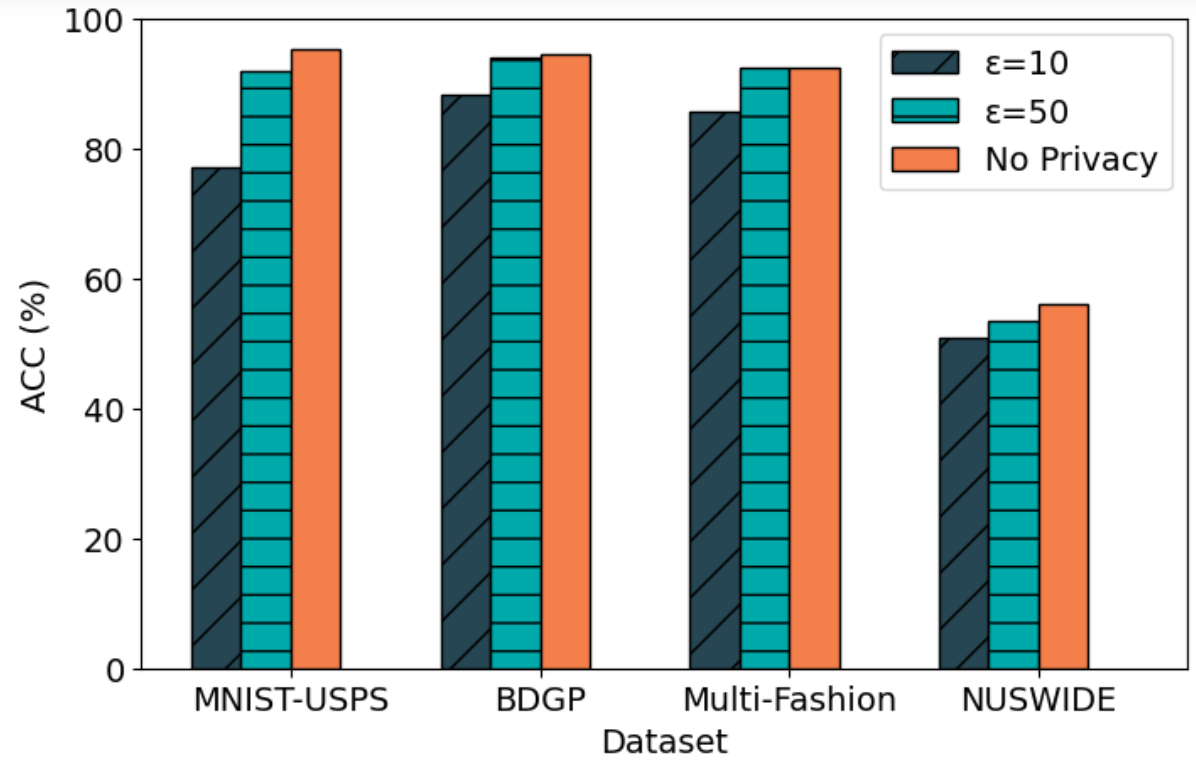}}
  \caption{(a) Effect of samples per client on generalization performance. (b) Scalability with the number of clients on Multi-Fashion. (c) Sensitivity under privacy constraints when $M/S$ = 2:1.}
  \label{fl}
\end{figure*}

\subsection{Attributes of Federated Learning}

\textbf{Generalization Analysis.} To check the validity of our theory for the proposed method, we investigate the impact of the number of samples per client and client participation rate on the generalization performance in the clustering task, as shown in Figure \ref{fl} (a) and Table \ref{gen}.
\begin{wraptable}{r}{0.7\textwidth}
  \caption{ Effect of participation rates on generalization performance.}
\centering
\begin{adjustbox}{width=0.7\textwidth}
\begin{tabular}{|c|r|c c c|c c c|}
\hline
\multicolumn{1}{|c}{} &
\multicolumn{1}{|r}{Client types} & \multicolumn{3}{|c}{Participating clients} & \multicolumn{3}{|c|}{Non-participating clients}        \\
\hline
\multicolumn{1}{|c}{} &
\multicolumn{1}{|r|}{Participation rate} & 50\% & 70\% & 90\% & 50\% & 70\% & 90\%    \\
\hline
\multirow{4}{*}{\rotatebox{90}{Dataset}} 
& MNIST-USPS &90.64	&91.71	&92.97	&89.11	&91.67	&92.06\\
& BDGP  &88.32	&89.48	&92.73	&85.92	&86.63	&87.65	\\
& Multi-Fashion &87.58	&89.46	&89.81	&87.31	&87.48	&88.73	\\
& NUSWIDE &53.12	&53.18	&53.58	&46.08	&51.46	&52.51 \\
\hline

\end{tabular}
\end{adjustbox}
  \label{gen}
\end{wraptable}
 In this setting, the total number of clients is fixed. We find that: i) Increasing the number of samples per client effectively raises the total number of samples involved in training, which enhances the model's generalization performance. ii) As the proportion of participating clients increases, the accuracy of non-participating clients also improves, and the performance gap between participating and non-participating clients narrows. These observations are consistent with the theoretical understanding in Theorem \ref{th2}, i.e., the generalization performance of the model is enhanced as both the number of samples per client and the number of participating clients increase.


\textbf{Number of Clients.}
We next consider the effects of changing the number of clients as shown in Figure \ref{fl} (b). It is observed that as the number of clients increases, the performance of FMCSC experiences a slight decline but remains generally stable. 
Only when the client number reaches 100, does a noticeable decline in performance occur, which is attributed to the insufficient number of samples within each client.

\textbf{Privacy.}
FMCSC, by design, does not share any raw data between clients and the server. Only the model parameters on each client are shared with the server. 
To further protect client privacy, we adopt differential privacy \cite{abadi2016deep} by adding noise to the model parameters uploaded from the client to the server.
Figure \ref{fl} (c) illustrates the clustering accuracy of FMCSC under different privacy bounds $\varepsilon$. We observe that FMCSC achieves both high performance and privacy at $\varepsilon$ = 50. However, as the level of noise increases at $\varepsilon$ = 10, the performance of FMCSC unavoidably degrades. 

\section{Conclusion}
In this paper, we propose FMCSC that can handle practical scenarios with heterogeneous hybrid views and explore the data cluster structures distributed on multiple clients.
First, we propose cross-client consensus pre-training to align the local models on all clients to avoid their misalignment.
Then, local-synergistic contrast and global-specific weighting aggregation are designed to bridge the client gap and the view gap across distributed clients and explore the cluster structures in multi-view data distributed on different clients.
Theoretical analysis and extensive experiments demonstrate that FMCSC outperforms state-of-the-art methods across diverse heterogeneous hybrid views and various federated learning scenarios.
In future work, we will use the method for more downstream tasks and some real-world situations, such as medical analysis and financial forecasting.


\bibliographystyle{plain}
\bibliography{references}


\newpage
\appendix
We provide more details and results about our work in the appendices. Here are the contents:
\begin{itemize}
    \item Appendix \ref{app:frame}: Framework of the proposed algorithm.
    \item Appendix \ref{app:work}: More related work.
    \item Appendix \ref{proof}: Proofs of Theorem 1 and Theorem 2.
    \item Appendix \ref{app:details}: More details about experimental settings.
    \item Appendix \ref{app:add}: Additional experiment results.
    \item Appendix \ref{app:impact}: Broader impacts of our proposed method.
    \item Appendix \ref{app:lim}: Limitations of our proposed method.
\end{itemize}

\section{Framework of the Proposed Algorithm} 
\label{app:frame}
Algorithm \ref{alg1} outlines the execution flow for both clients and the server in FMCSC. Initially, cross-client consensus pre-training aligns the unsupervised local models of each client (Lines 1-2). 
During the training, feature contrastive learning is employed for multi-view clients to explore common semantics among different views (Lines 5-12). For single-view clients, model contrastive learning is designed between local models and global models, promoting the extraction of more generalized common semantics from client models (Lines 13-19). 
Subsequently, the server develops global-specific weighting aggregation to aggregate multiple high-quality models (Lines 21-23). 
Finally, complementary cluster structures are discovered using the global models across all clients (Line 25).

\renewcommand{\algorithmicrequire}{\textbf{Input:}}  
\renewcommand{\algorithmicensure}{\textbf{Output:}} 
\begin{algorithm}[!h]
  \caption{Federated Multi-view Clustering via Synergistic Contrast (FMCSC)} 
  \label{alg1}
  \begin{algorithmic}[1]
    \Require
      Data with $V$ views distributed among $M$ multi-view clients and $S$ single-view clients, communication rounds $R$, Local epoch $E$.
    \Ensure
       Overall clustering results.
       \State Pre-train all client models by Eqs. (\ref{eq1})-(\ref{eq2}).
       \State server receives pre-trained models from all clients and initializes the global models.
       \While{not reaching $R$ rounds}
            \For{$c=1$ to $(M+S)$} \textbf{in parallel} 
                \If{$c \in$ $[M]$} \Comment{Multi-view clients}
                    \While{not reach the maximum iterations $E$}
                            \State Learn common semantics by Eqs. (\ref{eq4})-(\ref{eq5}).
                            \State Optimize the total loss by Eq. (\ref{eq8}).
                    \EndWhile
                    \State Optimize multiple global models by Eq. (\ref{eq6}).
                    \State Upload $\left\{f_{m}^v\left(\cdot; \mathbf{w}_m^{v}\right) \right\}_{v=1}^{V}$ and $ f_{m}\left(\cdot; \mathbf{w}_m\right)$ to the server.
                \ElsIf{$c \in$ $[S]$} \Comment{Single-view clients}
                    \While{not reach the maximum iterations $E$}
                            \State Learn common semantics by Eq. (\ref{eq7}).
                            \State Optimize the total loss by Eq. (\ref{eq8}).
                    \EndWhile
                    \State Upload $f_{p}\left(\cdot; \mathbf{w}_p^{v}\right)$ to the server.
                \EndIf           
            \EndFor             \Comment{Server}
        \State Assigning weights to local models by Eq. (\ref{eq9}). 
        \State Aggregate models among all clients by Eq. (\ref{eq10}).
        
        \State Distribute multiple global models to each client.
       \EndWhile
       \State Calculate the clustering results by Eq. (\ref{eq12}).
  \end{algorithmic}
\end{algorithm}

\section{More Related Work} \label{app:work}
Multi-view clustering (MVC) methods leverage consistency and complementary information between multiple views to enhance clustering effectiveness.  
    Based on their ability to handle missing data, existing MVC methods can be classified into two categories.
    Complete multi-view clustering \cite{huang2021non,ling2023dual,yan2023gcfagg,cui2023deep,huang2023self} which uncovers hidden patterns and structures by leveraging complete multi-view data for clustering.
    The success of existing complete multi-view clustering relies on the assumption of sample integrity across multiple views.
    However, in real-world scenarios, samples of multi-view are partially available due to data corruption or sensor failure, which leads to incomplete multi-view clustering studies \cite{liu2020efficient,yang2022robust,zhao2016incomplete,xu2023untie}.
    Incomplete multi-view clustering via prototype-based imputation (ProImp) \cite{li2023incomplete} employs a dual attention layer and a dual contrastive learning loss to learn view-specific prototypes and recover missing data.
    Cross-view partial sample and prototype alignment network (CPSPAN) \cite{jin2023deep} employs complete data to guide sample reconstruction and proposes shifted prototype alignment to calibrate prototype sets across views.
    However, the aforementioned MVC methods assume that multi-view data are stored within a single entity, thus lacking the concept of heterogeneous clients, and only addressing the hybrid views scenarios we mentioned.

\section{Theoretical Analysis} \label{proof}
\subsection{Proof of Theorem \ref{th1}}
    In this part, we want to prove that minimizing contrastive loss $\mathcal{L}_{c}^{m}$ and $\mathcal{L}_{c}^{p}$ are equal to maximizing mutual information. The proof is motivated by \cite{oord2018representation,zhong2020deep}.

    
\begin{proof} 
    $\mathcal{L}_{c}^{m}$ and $\mathcal{L}_{c}^{p}$ as the contrastive loss, denoting the consistency objectives of the multi-view client $m$ and single-view client $p$ respectively, i.e.,
    
    $$\mathcal{L}_{c}^{m} = -\frac{1}{{\left|\mathcal{M}_{m}\right|}} \sum_{v=1}^{V}\sum_{i=1}^{{\left|\mathcal{M}_{m}\right|}} \log \frac{e^{s i m\left(\mathbf{h}_{i}, \mathbf{h}_{i}^{v}\right) / \tau_m}}{\sum_{j\ne i}  e^{sim\left(\mathbf{h}_{i}, \mathbf{h}_{j}^{v}\right) / \tau_m}},$$

    $$\mathcal{L}_{c}^{p} = -\frac{1}{{\left|\mathcal{S}_{p}\right|}} \sum_{i=1}^{{\left|\mathcal{S}_{p}\right|}} \log \frac{e^{s i m\left(\mathbf{h}_{i}, \mathbf{h}_{i}^{g}\right) / \tau_p}}{e^{sim\left(\mathbf{h}_{i}, \mathbf{h}_{i}^{g}\right) / \tau_p}+e^{sim\left(\mathbf{h}_{i}, \mathbf{z}_{i}^{v}\right) / \tau_p}}.$$
    
    When $j \ne i$, we assume that $p\left(\mathbf{h}_{i}, \mathbf{h}_{j}^{v}\right)=p\left(\mathbf{h}_{i}\right) p\left(\mathbf{h}_{j}^{v}\right)$, which means that $\frac{p\left(\mathbf{h}_{i}, \mathbf{h}_{j}^{v}\right)}{p\left(\mathbf{h}_{i}\right) p\left(\mathbf{h}_{j}^{v}\right)}=1$.
    Let $\mathcal{N}_{i} = \sum_{j=1}^{\left|\mathcal{M}_{m}\right|} \frac{p\left(\mathrm{h}_{i}, \mathrm{h}_{j}^{v}\right)}{p\left(\mathrm{h}_{i}\right) p\left(\mathrm{h}_{j}^{v}\right)}$, we can then

    $$ \begin{aligned}
    I\left(\mathbf{H} ; \mathbf{H}^{v}\right) & =\sum_{i=1}^{{\left|\mathcal{M}_{m}\right|}} \sum_{j=1}^{{\left|\mathcal{M}_{m}\right|}} p\left(\mathbf{h}_{i}, \mathbf{h}_{j}^{v}\right) \log \frac{p\left(\mathbf{h}_{i}, \mathbf{h}_{j}^{v}\right)}{p\left(\mathbf{h}_{i}\right) p\left(\mathbf{h}_{j}^{v}\right)} \\
    & =\sum_{i=1}^{{\left|\mathcal{M}_{m}\right|}} p\left(\mathbf{h}_{i}, \mathbf{h}_{i}^{v}\right) \log \frac{p\left(\mathbf{h}_{i}, \mathbf{h}_{i}^{v}\right)}{p\left(\mathbf{h}_{i}\right) p\left(\mathbf{h}_{i}^{v}\right)}+
    \sum_{i=1}^{{\left|\mathcal{M}_{m}\right|}} \sum_{j\ne i} p\left(\mathbf{h}_{i}, \mathbf{h}_{j}^{v}\right) \log \frac{p\left(\mathbf{h}_{i}, \mathbf{h}_{j}^{v}\right)}{p\left(\mathbf{h}_{i}\right) p\left(\mathbf{h}_{j}^{v}\right)} \\
    & =\sum_{i=1}^{{\left|\mathcal{M}_{m}\right|}} p\left(\mathbf{h}_{i}, \mathbf{h}_{i}^{v}\right) \log \left(\frac{p\left(\mathbf{h}_{i}, \mathbf{h}_{i}^{v}\right)}{p\left(\mathbf{h}_{i}\right) p\left(\mathbf{h}_{i}^{v}\right) \cdot \mathcal{N}_{i}} \cdot \mathcal{N}_{i}\right) \\
    & =\sum_{i=1}^{{\left|\mathcal{M}_{m}\right|}} p\left(\mathbf{h}_{i}, \mathbf{h}_{i}^{v}\right) \log \frac{\frac{p\left(\mathbf{h}_{i}, \mathbf{h}_{i}^{v}\right)}{p\left(\mathbf{h}_{i}\right) p\left(\mathbf{h}_{i}^{v}\right)}}{\mathcal{N}_{i}}+\sum_{i=1}^{{\left|\mathcal{M}_{m}\right|}} p\left(\mathbf{h}_{i}, \mathbf{h}_{i}^{v}\right) \log \mathcal{N}_{i}.
    \end{aligned}$$

    Since positive pairs are correlated, we have the estimate: $p\left(\mathbf{h}_{i}, \mathbf{h}_{i}^{v}\right) \geq p\left(\mathbf{h}_{i}\right) p\left(\mathbf{h}_{i}^{v}\right)$. In addition, we assume that there exists a constant $\delta_m \in(0,1)$ such that $p\left(\mathbf{h}_{i}^v \mid \mathbf{h}_{i} \right)>\delta_m$, $i=1,2, \cdots, \left|\mathcal{M}_{m}\right| $ holds. According to \cite{xu2023self} and with the estimation, we have $p\left(\mathbf{h}_{i}\right) \approx \frac{1}{\left|\mathcal{M}_{m}\right|}, i=1,2, \cdots, \left|\mathcal{M}_{m}\right|$, and $e^{ sim\left(\mathbf{h}_{i}, \mathbf{h}_{j}^{v}\right) / \tau_m} \propto \frac{p\left(\mathbf{h}_{i}, \mathbf{h}_{j}^{v}\right)}{p\left(\mathbf{h}_{i}\right) p\left(\mathbf{h}_{j}^{v}\right)}$, the following inequality holds:

    $$\begin{aligned}
    \sum_{v=1}^{V}I\left(\mathbf{H}, \mathbf{H}^{v}\right) & =\sum_{v=1}^{V}\sum_{i=1}^{{\left|\mathcal{M}_{m}\right|}} p\left(\mathbf{h}_{i}, \mathbf{h}_{i}^{v}\right) \log \frac{\frac{p\left(\mathbf{h}_{i}, \mathbf{h}_{i}^{v}\right)}{p\left(\mathbf{h}_{i}\right) p\left(\mathbf{h}_{i}^{v}\right)}}{\mathcal{N}_{i}}+
    \sum_{v=1}^{V}\sum_{i=1}^{{\left|\mathcal{M}_{m}\right|}} p\left(\mathbf{h}_{i}, \mathbf{h}_{i}^{v}\right) \log \left( \sum_{j=1}^{\left|\mathcal{M}_{m}\right|} \frac{p\left(\mathbf{h}_{i}, \mathbf{h}_{j}^{v}\right)}{p\left(\mathbf{h}_{i}\right) p\left(\mathbf{h}_{j}^{v}\right)}\right) \\
    & \approx \sum_{v=1}^{V}\sum_{i=1}^{\left|\mathcal{M}_{m}\right|} \frac{1}{\left|\mathcal{M}_{m}\right|} p\left(\mathbf{h}_{i}^{v} \mid \mathbf{h}_{i}\right) \log \frac{\frac{p\left(\mathbf{h}_{i}, \mathbf{h}_{i}^{v}\right)}{p\left(\mathbf{h}_{i}\right) p\left(\mathbf{h}_{i}^{v}\right)}}{\mathcal{N}_{i}}
    +\sum_{v=1}^{V} \log \left(\left|\mathcal{M}_{m}\right|-1+\frac{p\left(\mathbf{h}_{i}, \mathbf{h}_{i}^{v}\right)}{p\left(\mathbf{h}_{i}\right) p\left(\mathbf{h}_{i}^{v}\right)}\right) \\
    & \geq \frac{\delta_m}{\left|\mathcal{M}_{m}\right|} \sum_{v=1}^{V} \sum_{i=1}^{\left|\mathcal{M}_{m}\right|} \log \frac{e^{ sim\left(\mathbf{h}_{i}, \mathbf{h}_{i}^{v}\right) / \tau_m}}{\sum_{j\ne i}  e^{\operatorname{sim}\left(\mathbf{h}_{i}, \mathbf{h}_{j}^{v}\right) / \tau_m} + e^{ sim\left(\mathbf{h}_{i}, \mathbf{h}_{i}^{v}\right) / \tau_m} }+V \log \left|\mathcal{M}_{m}\right| \\
    & \geq V \log \left|\mathcal{M}_{m}\right|-\delta_m \mathcal{L}_{c}^{m}.
    \end{aligned}$$

    Similarly,  we assume that there exists a constant $\delta_p
    \in(0,1)$ such that $p\left(\mathbf{h}_{i}^g \mid \mathbf{h}_{i} \right)=p\left(\mathbf{z}_{i}^v \mid \mathbf{h}_{i} \right)>\delta_p$, $i=1,2, \cdots, \left|\mathcal{S}_{p}\right| $ holds. And we have $p\left(\mathbf{h}_{i}\right) \approx \frac{1}{\left|\mathcal{S}_{p}\right|}$, $e^{ sim\left(\mathbf{h}_{i}, \mathbf{h}_{i}^{g}\right) / \tau_p} \propto \frac{p\left(\mathbf{h}_{i}, \mathbf{h}_{i}^{g}\right)}{p\left(\mathbf{h}_{i}\right) p\left(\mathbf{h}_{i}^{g}\right)}$ and $e^{ sim\left(\mathbf{h}_{i}, \mathbf{z}_{i}^{v}\right) / \tau_p} \propto \frac{p\left(\mathbf{h}_{i}, \mathbf{z}_{i}^{v}\right)}{p\left(\mathbf{h}_{i}\right) p\left(\mathbf{z}_{i}^{v}\right)}$, the following inequality holds:
    
    $$\begin{aligned}
    I\left(\mathbf{H}, \mathbf{H}^{g}\right)-I\left(\mathbf{H}, \mathbf{Z}^{v}\right) = & \sum_{i=1}^{{\left|\mathcal{S}_{p}\right|}} p\left(\mathbf{h}_{i}, \mathbf{h}_{i}^{g}\right) \log \frac{p\left(\mathbf{h}_{i}, \mathbf{h}_{i}^{g}\right)}{p\left(\mathbf{h}_{i}\right) p\left(\mathbf{h}_{i}^{g}\right)}-\sum_{i=1}^{{\left|\mathcal{S}_{p}\right|}} p\left(\mathbf{h}_{i}, \mathbf{z}_{i}^{v}\right) \log \frac{p\left(\mathbf{h}_{i}, \mathbf{z}_{i}^{v}\right)}{p\left(\mathbf{h}_{i}\right) p\left(\mathbf{z}_{i}^{v}\right)} \\
    \approx &  \sum_{i=1}^{\left|\mathcal{S}_{p}\right|} \frac{2}{\left|\mathcal{S}_{p}\right|} p\left(\mathbf{h}_{i}^{g} \mid \mathbf{h}_{i}\right) \log \frac{p\left(\mathbf{h}_{i}, \mathbf{h}_{i}^{g}\right)}{p\left(\mathbf{h}_{i}\right) p\left(\mathbf{h}_{i}^{g}\right)}
    - \sum_{i=1}^{\left|\mathcal{S}_{p}\right|} \frac{1}{\left|\mathcal{S}_{p}\right|} p\left(\mathbf{h}_{i}^{g} \mid \mathbf{h}_{i}\right) \log \frac{p\left(\mathbf{h}_{i}, \mathbf{h}_{i}^{g}\right)}{p\left(\mathbf{h}_{i}\right) p\left(\mathbf{h}_{i}^{g}\right)} \\
    & -\sum_{i=1}^{\left|\mathcal{S}_{p}\right|} \frac{1}{\left|\mathcal{S}_{p}\right|} p\left(\mathbf{z}_{i}^{v} \mid \mathbf{h}_{i}\right) \log \frac{p\left(\mathbf{h}_{i}, \mathbf{z}_{i}^{v}\right)}{p\left(\mathbf{h}_{i}\right) p\left(\mathbf{z}_{i}^{v}\right)} \\ 
    \leq & \frac{2\delta_p}{\left|\mathcal{S}_{p}\right|} \sum_{i=1}^{\left|\mathcal{S}_{p}\right|} \log \frac{p\left(\mathbf{h}_{i}, \mathbf{h}_{i}^{g}\right)}{p\left(\mathbf{h}_{i}\right) p\left(\mathbf{h}_{i}^{g}\right)}
    -\frac{2\delta_p}{\left|\mathcal{S}_{p}\right|} \sum_{i=1}^{\left|\mathcal{S}_{p}\right|} \log \left( \frac{p\left(\mathbf{h}_{i}, \mathbf{h}_{i}^{g}\right)}{p\left(\mathbf{h}_{i}\right) p\left(\mathbf{h}_{i}^{g}\right)}+\frac{p\left(\mathbf{h}_{i}, \mathbf{z}_{i}^{v}\right)}{p\left(\mathbf{h}_{i}\right) p\left(\mathbf{z}_{i}^{v}\right)}\right) \\
    \leq & \frac{2\delta_p}{\left|\mathcal{S}_{p}\right|} \sum_{i=1}^{\left|\mathcal{S}_{p}\right|}\log \frac{e^{s i m\left(\mathbf{h}_{i}, \mathbf{h}_{i}^{g}\right) / \tau_p}}{e^{sim\left(\mathbf{h}_{i}, \mathbf{h}_{i}^{g}\right) / \tau_p}+e^{sim\left(\mathbf{h}_{i}, \mathbf{z}_{i}^{v}\right) / \tau_p}} \\
    \leq & -2\delta_p \mathcal{L}_{c}^{p}.
    \end{aligned}$$
\end{proof}

\subsection{Proof of Theorem \ref{th2}}
We consider a heterogeneous federated learning setting where $M$ multi-view clients and $S$ single-view clients. For each multi-view client, we define the empirical risk and its expectation as $\widehat{\mathcal{L}}_m(f)$ and $\mathcal{L}_m(f)$, respectively. Similarly, for each single-view client, the empirical risk and its expectation are denoted as $\widehat{\mathcal{L}}_p(f)$ and $\mathcal{L}_p(f)$. 
Let $f: \mathcal{X} \rightarrow \mathbb{R}^{D_{v}} \times \mathbb{R}^{d_{v}} \times \mathbb{R}^{d_{v}}$ denotes the function that maps input samples into reconstruction samples, low-level features and high-level features. Then, the reconstruction samples, low-level features and high-level features are given by $\hat{\mathbf{x}}_{i}^{v}:=f_{x}\left(\mathbf{x}_{i}^{v}\right) \in \mathbb{R}^{D_{v}}$, 
$\mathbf{z}_i^v:=f_{z}^v\left( \mathbf{x}_{i}^{v}\right) \in \mathbb{R}^{d_v}$, and
$\mathbf{h}_i^v:=f_{h}^v\left( \mathbf{x}_{i}^{v}\right) \in \mathbb{R}^{d_v}$, respectively.
Additionally, we define $\left \{ \mathbf{x}_{i}^{v} \right \} _{v=1}^V := \mathbf{x}_{i}$, then common semantics are denoted as
$\mathbf{h}_i:=f_{h}^0\left(\mathbf{x}_{i}\right) \in \mathbb{R}^{d}$. To prove Theorem \ref{th2}, we first introduce the following three lemmas.

\begin{lemma}\label{lemma1}
    We define the empirical risk $\widehat{\mathcal{L} }_m(f)$ for multi-view client $m$ as
    $$\begin{aligned}
    \widehat{\mathcal{L} }_m(f) &=\frac{1}{\left| \mathcal{M} _m \right|}\sum_{v=1}^V{\sum_{i=1}^{\left| \mathcal{M} _m \right|}{\left[ \left\| \mathbf{x}_{i}^{v}-f_x\left( \mathbf{x}_{i}^{v} \right) \right\| ^2-\log \frac{e^{sim\left( \mathbf{h}_i,\mathbf{h}_{i}^{v} \right) /\tau _m}}{\sum_{j\ne i}{e^{sim\left( \mathbf{h}_i,\mathbf{h}_{j}^{v} \right) /\tau _m}}} \right]}}
    \\ &
    =\frac{1}{\left| \mathcal{M} _m \right|}\sum_{v=1}^V{\sum_{i=1}^{\left| \mathcal{M} _m \right|}{\left[ \left\| \mathbf{x}_{i}^{v}-f_x\left( \mathbf{x}_{i}^{v} \right) \right\| ^2-\log \frac{e^{sim\left( f_{h}^0\left(\mathbf{x}_{i}\right),f_{h}^v\left( \mathbf{x}_{i}^{v}\right) \right) /\tau _m}}{\sum_{j\ne i}{e^{sim\left( f_{h}^0\left(\mathbf{x}_{i}\right),f_{h}^v\left( \mathbf{x}_{j}^{v}\right) \right) /\tau _m}}} \right]}}.
    \end{aligned}$$
    
    Let $\mathcal{L}_m(f)$ be the expectation of $\widehat{\mathcal{L} }_m(f)$. Suppose that for any $\mathbf{x} \in \mathcal{X}$ and $f \in \mathcal{F}$, there exists $D<\infty$ such that $\left\|\mathbf{x}\right\|,\left\|f_x\left(\mathbf{x}\right)\right\|,\left\|f_{h}^v\left(\mathbf{x}\right)\right\|,\left\|f_{h}^0\left(\mathbf{x}\right)\right\| \in[0, D]$ hold.
    With probability $1-\delta$ for any $f \in \mathcal{F}$, the following inequality holds
    $$\mathcal{L}_m(f) \leq \widehat{\mathcal{L}}_m(f)+\frac{12V D^{2}}{\sqrt{\left| \mathcal{M} _m \right|}}+9V D^{2} \sqrt{\frac{\log \frac{1}{\delta}}{2 \left| \mathcal{M} _m \right|}} .$$
\end{lemma}

\begin{proof} 
This proof is inspired by \cite{tang2022deep}. For simplicity, we define $f_{h}\left(\mathbf{x}_i,\mathbf{x}_j\right):=sim\left( f_{h}^0\left(\mathbf{x}_{i}\right),f_{h}^v\left( \mathbf{x}_{j}^{v}\right) \right) /\tau _m$ and $f_{h}\left(\mathbf{x}_i\right):=sim\left( f_{h}^0\left(\mathbf{x}_{i}\right),f_{h}^v\left( \mathbf{x}_{i}^{v}\right) \right) /\tau _m$, respectively. Then, the empirical risk and its expectation can be formulated as

$$\begin{aligned}
\widehat{\mathcal{L} }_m(f) & =\frac{1}{\left| \mathcal{M} _m \right|}\sum_{v=1}^V{\sum_{i=1}^{\left| \mathcal{M} _m \right|}{\left[ \left\| \mathbf{x}_{i}^{v}-f_x\left( \mathbf{x}_{i}^{v} \right) \right\| ^2-\log \frac{e^{f_{h}\left(\mathbf{x}_i\right)}}{\sum_{j\ne i}{e^{f_{h}\left(\mathbf{x}_i,\mathbf{x}_{j}\right)}}} \right]}}
\\ &
=\frac{1}{\left| \mathcal{M} _m \right|}\sum_{v=1}^V{\sum_{i=1}^{\left| \mathcal{M} _m \right|}{\left[ \left\| \mathbf{x}_{i}^{v}-f_x\left( \mathbf{x}_{i}^{v} \right) \right\| ^2+\log\left( \sum_{j\ne i}{\exp \left\{ f_{h}\left(\mathbf{x}_i,\mathbf{x}_{j}\right) \right\}}\right)-f_{h}\left(\mathbf{x}_i\right) \right]}}
\\ &
=\frac{1}{\left| \mathcal{M} _m \right|}\sum_{v=1}^V{\sum_{i=1}^{\left| \mathcal{M} _m \right|}{\left[ \left\| \mathbf{x}_{i}^{v}-f_x\left( \mathbf{x}_{i}^{v} \right) \right\| ^2-f_{h}\left(\mathbf{x}_i\right) \right]}}
+\frac{1}{\left| \mathcal{M} _m \right|}\sum_{v=1}^V{\sum_{i=1}^{\left| \mathcal{M} _m \right|}{\log \sum_{j\ne i}{\exp \left\{ f_{h}\left(\mathbf{x}_i,\mathbf{x}_{j}\right) \right\}}}},
\end{aligned}
$$
and
$$\begin{aligned}
\mathcal{L}_m(f) & =\sum_{v=1}^V{\mathbb{E}_\mathbf{x}\left[ \left\| x_{i}^{v}-f_x\left( x_{i}^{v} \right) \right\| ^2-f_{h}\left(\mathbf{x}_i\right) \right]}
+\sum_{v=1}^V{\mathbb{E}_\mathbf{x}\left[\log \sum_{j\ne i}{\exp \left\{ f_{h}\left(\mathbf{x}_i,\mathbf{x}_{j}\right) \right\}}\right]}.
\end{aligned} $$

Let $\bar{\mathcal{M}} _m$ be the sample set that different from $\mathcal{M} _m$ by only one set of data $ \bar{\mathbf{x}}_{r} := \left \{ \bar{\mathbf{x}}_{r}^{v} \right \} _{v=1}^V$. The empirical risk of the function $f$ on $\bar{\mathcal{M}} _m$ is denoted as $\widehat{\mathcal{L}}^{\prime}_m(f)$. We have

$$\begin{aligned}
& \left|\sup _{f \in \mathcal{F}}| \mathcal{L}_m(f)-\widehat{\mathcal{L}}_m(f)|-\sup _{f \in \mathcal{F}}| \mathcal{L}_m(f)-\widehat{\mathcal{L}}^{\prime}_m(f)|\right| \\
\leq & \sup _{f \in \mathcal{F}}\left|\widehat{\mathcal{L}}_m(f)-\widehat{\mathcal{L}}^{\prime}_m(f)\right| \\
\leq & \sup _{f \in \mathcal{F}}\left|\frac{1}{\left| \mathcal{M} _m \right|}\sum_{v=1}^V{\left(\left\|\mathbf{x}_{r}^{v}-f_x\left(\mathbf{x}_{r}^{v}\right)\right\|^{2}-\left\|\bar{\mathbf{x}}_{r}^{v}-f_x\left(\bar{\mathbf{x}}_{r}^{v}\right)\right\|^{2}
-\left(f_{h}\left(\mathbf{x}_{r}\right)-f_{h}\left(\bar{\mathbf{x}}_{r}\right)\right)\right)}\right| \\
& + \sup _{f \in \mathcal{F}}\left|\frac{2}{\left| \mathcal{M} _m \right|}\sum_{v=1}^V{\left(\log \sum_{j\ne i}{\exp \left\{ f_{h}\left(\mathbf{x}_r,\mathbf{x}_{j}\right) \right\}}-\log \sum_{j\ne i}{\exp \left\{ f_{h}\left(\bar{\mathbf{x}}_r,\mathbf{x}_{j}\right) \right\}}\right)}\right|
\\
\leq & \sup _{f \in \mathcal{F}} \frac{1}{\left| \mathcal{M} _m \right|}\sum_{v=1}^V{\left|\left\|\mathbf{x}_{r}^{v}\right\|^{2}-\left\|\bar{\mathbf{x}}_{r}^{v}\right\|^{2}+\left\|f_x\left(\mathbf{x}_{r}^{v}\right)\right\|^{2}-\left\|f_x\left(\bar{\mathbf{x}}_{r}^{v}\right)\right\|^{2}+2\left\|\mathbf{x}_{r}^{v}\right\|\left\|f_x\left(\mathbf{x}_{r}^{v}\right)\right\|+2\left\|\bar{\mathbf{x}}_{r}^{v}\right\|\left\|f_x\left(\bar{\mathbf{x}}_{r}^{v}\right)\right\|\right|}
\\
& + \sup _{f \in \mathcal{F}} \frac{1}{\left| \mathcal{M} _m \right|}\sum_{v=1}^V{\left| f_{h}\left(\mathbf{x}_{r}\right)-f_{h}\left(\bar{\mathbf{x}}_{r}\right)\right|}
+ \sup _{f \in \mathcal{F}} \frac{2}{\left| \mathcal{M} _m \right|\left(\left| \mathcal{M} _m \right|-1\right)}\sum_{v=1}^V{\left|\sum_{j\ne i}{\left(f_{h}\left(\mathbf{x}_{r},\mathbf{x}_{j}\right)-f_{h}\left(\bar{\mathbf{x}}_{r},\mathbf{x}_{j}\right)\right)} \right|}
\\
\leq & \frac{9 VD^{2}}{\left| \mathcal{M} _m \right|}.
\end{aligned} $$

Then, we analyze the upper bound of the expectation term $\mathbb{E} \sup _{f \in \mathcal{F}}|\widehat{\mathcal{L}}_m(f)-\mathcal{L}_m(f)|$. Let $\sigma_{1}, \ldots, \sigma_{\left| \mathcal{M} _m \right|}$ be i.i.d. independent random variables taking values in $\{-1,1\}$ and $\bar{\mathcal{M}} _m$ be the independent copy of $\mathcal{M} _m$. We have

$$\begin{aligned}
& \mathbb{E} \sup _{f \in \mathcal{F}}|\widehat{\mathcal{L}}_m(f)-\mathcal{L}_m(f)| \\
\leq & \mathbb{E}_{\mathcal{M} _m, \bar{\mathcal{M}} _m} \sup _{f \in \mathcal{F}}\left|\frac{1}{\left| \mathcal{M} _m \right|}\sum_{v=1}^V{\sum_{i=1}^{\left| \mathcal{M} _m \right|}{\left(\left\|\mathbf{x}_{i}^{v}-f_x\left(\mathbf{x}_{i}^{v}\right)\right\|^{2}-\left\|\bar{\mathbf{x}}_{i}^{v}-f_x\left(\bar{\mathbf{x}}_{i}^{v}\right)\right\|^{2}\right)}}\right|
\\
&+  \mathbb{E}_{\mathcal{M} _m, \bar{\mathcal{M}} _m} \sup _{f \in \mathcal{F}}\left|\frac{1}{\left| \mathcal{M} _m \right|}\sum_{v=1}^V{\sum_{i=1}^{\left| \mathcal{M} _m \right|}{\left(f_{h}\left(\mathbf{x}_{i}\right)-f_{h}\left(\bar{\mathbf{x}}_{i}\right)\right)}}\right|
\\
& +  \mathbb{E}_{\mathcal{M} _m, \bar{\mathcal{M}} _m} \sup _{f \in \mathcal{F}}\left|\frac{2}{\left| \mathcal{M} _m \right|}\sum_{v=1}^V{\sum_{i=1}^{\left| \mathcal{M} _m \right|}{\left(\log \sum_{j\ne i}{\exp \left\{ f_{h}\left(\mathbf{x}_i,\mathbf{x}_{j}\right) \right\}}-\log \sum_{j\ne i}{\exp \left\{ f_{h}\left(\bar{\mathbf{x}}_i,\mathbf{x}_{j}\right) \right\}}
\right)}}\right|
\\
\leq & 2\mathbb{E}_{\mathcal{M} _m, \sigma} \sup _{f \in \mathcal{F}} \left\lvert\, \frac{1}{\left| \mathcal{M} _m \right|}\sum_{v=1}^{V} \sum_{i=1}^{\left| \mathcal{M} _m \right|} \sigma_{i}\left\|\mathbf{x}_{i}^{v}-f_x\left(\mathbf{x}_{i}^{v}\right)\right\|^{2}\right|
+ 2 \mathbb{E}_{\mathcal{M} _m, \sigma} \sup _{f \in \mathcal{F}}\left|\frac{1}{\left| \mathcal{M} _m \right|}\sum_{v=1}^V{\sum_{i=1}^{\left| \mathcal{M} _m \right|}{\sigma_{i} f_{h}\left(\mathbf{x}_{i}\right)}}\right|
\\
& +  2 \mathbb{E}_{\mathcal{M} _m, \sigma} \sup _{f \in \mathcal{F}}\left|\frac{2}{\left| \mathcal{M} _m \right|\left(\left| \mathcal{M} _m \right|-1\right)}\sum_{v=1}^V{\sum_{i=1}^{\left| \mathcal{M} _m \right|}{ \sum_{j\ne i}\sigma_{i}{f_{h}\left(\mathbf{x}_i,\mathbf{x}_{j}\right) }}}\right|
\\
\leq & 2 V \max _{v} \mathbb{E}_{\mathcal{M} _m, \sigma} \sup _{f \in \mathcal{F}}\left(\frac{1}{\left| \mathcal{M} _m \right|} \sum_{i=1}^{\left| \mathcal{M} _m \right|}\left[\left\|\mathbf{x}_{i}^{v}-f_x\left(\mathbf{x}_{i}^{v}\right)\right\|^{2}\right]^{2}\right)^{\frac{1}{2}}
+ 2 V \max _{v} \mathbb{E}_{\mathcal{M} _m, \sigma} \sup _{f \in \mathcal{F}}\left(\frac{1}{\left| \mathcal{M} _m \right|} \sum_{i=1}^{\left| \mathcal{M} _m \right|}\left[f_{h}\left(\mathbf{x}_{i}\right)\right]^{2}\right)^{\frac{1}{2}}
\\
& + 4 V \max _{v} \mathbb{E}_{\mathcal{M} _m, \sigma} \sup _{f \in \mathcal{F}}\left(\frac{1}{\left| \mathcal{M} _m \right|\left(\left| \mathcal{M} _m \right|-1\right)} \sum_{i=1}^{\left| \mathcal{M} _m \right|}{ \sum_{j\ne i}}\left[f_{h}\left(\mathbf{x}_{i}\right)\right]^{2}\right)^{\frac{1}{2}},
\\
\leq & \frac{12 VD^{2}}{\sqrt{\left| \mathcal{M} _m \right|}},
\end{aligned}$$

where the second last inequality inequality is obtained by the Khintchine-Kahane inequality\cite{latala1994best}. Thus, according to the McDiarmid inequality \cite{mohri2018foundations}, with probability at least $1-\delta$ for any $f \in \mathcal{F}$, we have

$$\mathcal{L}_m(f) \leq \widehat{\mathcal{L}}_m(f)+\frac{12V D^{2}}{\sqrt{\left| \mathcal{M} _m \right|}}+9V D^{2} \sqrt{\frac{\log \frac{1}{\delta}}{2 \left| \mathcal{M} _m \right|}} .$$
\end{proof}

Due to the presence of view gaps among different views, let the data from the $v$-th view follow the distribution $\mathcal{D}_v$, and the multi-view data follow the distribution $\mathcal{D}$. We analyze the generalization bounds for $f_{m}^v\left(\cdot; \mathbf{w}_m^{v}\right)$ in multi-view client $m$, which is built upon prior works from domain adaptation.
\begin{lemma}[Generalization Bounds for Domain Adaptation \cite{ben2010theory,ben2006analysis}]\label{lemma2}
Consider the $v$-th view data domain $\mathcal{D}_v$ and the multi-view data domain $\mathcal{D}$, respectively. Given a feature extraction function $\mathcal{R}: \mathcal{X} \mapsto \mathcal{H}$ that shared between $\mathcal{D}_v$ and $\mathcal{D}$. Let $\mathcal{F}$ be a set of hypothesis with VC-dimension $d$. Then, for every $f \in \mathcal{F}$, with probability at least $1-\delta$:
    $$\mathcal{L}_{m}(f) \leq \mathcal{L}_{m^v}(f)+\sqrt{\frac{4}{\left| \mathcal{M} _m \right|}\left(d \log \frac{2 e \left| \mathcal{M} _m \right|}{d}+\log \frac{4}{\delta}\right)}+d_{\mathcal{F}}\left(\tilde{\mathcal{D}}_{v}, \tilde{\mathcal{D}}\right)+\lambda_{v},$$
where $d_{\mathcal{F}}\left(\tilde{\mathcal{D}}_{v}, \tilde{\mathcal{D}}\right)$ denotes the divergence measured over a symmetric-difference hypothesis space. $\tilde{\mathcal{D}}_{v}$ and $\tilde{\mathcal{D}}$ are the induced distributions of $\mathcal{D}_v$ and $\mathcal{D}$ under $\mathcal{R}$, respectively, s.t. $\mathbb{E}_{h \sim \tilde{\mathcal{D}}_{v}}[\mathcal{B}(h)]=\mathbb{E}_{x \sim \mathcal{D}_{v}}[\mathcal{B}(\mathcal{R}(x))]$ given a probability event $\mathcal{B}$, and so for $\tilde{\mathcal{D}}$. $\lambda_{v}:=\min _{f} \mathcal{L}_{m^v}(f)+\mathcal{L}_{m}(f)$ denotes an oracle performance.
\end{lemma}

\begin{lemma}\label{lemma3}
    We define the empirical risk $\widehat{\mathcal{L} }_p(f)$ for single-view client $p$ as
    $$\begin{aligned}
    \widehat{\mathcal{L} }_p(f) &=\frac{1}{\left| \mathcal{S} _p \right|}\sum_{i=1}^{\left| \mathcal{S} _p \right|}{\left[ \left\| \mathbf{x}_{i}^{v}-f_x\left( \mathbf{x}_{i}^{v} \right) \right\| ^2-\log \frac{e^{sim\left( \mathbf{h}_i,\mathbf{h}_{i}^{g} \right) /\tau _p}}{{e^{sim\left( \mathbf{h}_i,\mathbf{h}_{i}^{g} \right) /\tau _p}}+{e^{sim\left( \mathbf{h}_i,\mathbf{z}_{i}^{v} \right) /\tau _p}}} \right]}
    \\ &
    =\frac{1}{\left| \mathcal{S} _p \right|}\sum_{i=1}^{\left| \mathcal{S} _p \right|}{\left[ \left\| \mathbf{x}_{i}^{v}-f_x\left( \mathbf{x}_{i}^{v} \right) \right\| ^2-\log \frac{e^{sim\left( f_{h}^0\left(\mathbf{x}_{i}\right),f_{g}^0\left( \mathbf{x}_{i}\right) \right) /\tau _p}}{{e^{sim\left( f_{h}^0\left(\mathbf{x}_{i}\right),f_{g}^0\left( \mathbf{x}_{i}\right) \right) /\tau _p}+e^{sim\left( f_{h}^0\left(\mathbf{x}_{i}\right),f_{z}^v\left( \mathbf{x}_{i}^{v}\right) \right) /\tau _p}}} \right]}.
    \end{aligned}$$
    
    Let $\mathcal{L}_p(f)$ be the expectation of $\widehat{\mathcal{L} }_p(f)$. Suppose that for any $\mathbf{x} \in \mathcal{X}$ and $f \in \mathcal{F}$, there exists $D<\infty$ such that $\left\|\mathbf{x}\right\|,\left\|f_{x}\left(\mathbf{x}\right)\right\|,\left\|f_{z}^v\left(\mathbf{x}\right)\right\|,\left\|f_{h}^0\left(\mathbf{x}\right)\right\|, \left\|f_{g}^0\left(\mathbf{x}\right)\right\| \in[0, D]$ hold.
    With probability $1-\delta$ for any $f \in \mathcal{F}$, the following inequality holds
    $$\mathcal{L}_p(f) \leq \widehat{\mathcal{L}}_p(f)+\frac{10 D^{2}}{\sqrt{\left| \mathcal{S} _p \right|}}+8 D^{2} \sqrt{\frac{\log \frac{1}{\delta}}{2 \left| \mathcal{S} _p \right|}} .$$
\end{lemma}

\begin{proof} 
For simplicity, we define $f_{g}\left(\mathbf{x}_i\right):=sim\left( f_{h}^0\left(\mathbf{x}_{i}\right),f_{g}^0\left( \mathbf{x}_{i}\right) \right) /\tau _p$ and $f_{h,z}\left(\mathbf{x}_i\right):=sim\left( f_{h}^0\left(\mathbf{x}_{i}\right),f_{z}^v\left( \mathbf{x}_{i}^{v}\right) \right) /\tau _p$, respectively. Then, the empirical risk and its expectation can be formulated as

$$\begin{aligned}
\widehat{\mathcal{L} }_p(f) & =\frac{1}{\left| \mathcal{S} _p \right|}\sum_{i=1}^{\left| \mathcal{S} _p \right|}{\left[ \left\| \mathbf{x}_{i}^{v}-f_x\left( \mathbf{x}_{i}^{v} \right) \right\| ^2-\log \frac{e^{f_{g}\left(\mathbf{x}_i\right)}}{{e^{f_{g}\left(\mathbf{x}_i\right)}+e^{f_{h,z}\left(\mathbf{x}_i\right)}}} \right]}
\\ &
=\frac{1}{\left| \mathcal{S} _p \right|}\sum_{i=1}^{\left| \mathcal{S} _p \right|}{\left[ \left\| \mathbf{x}_{i}^{v}-f_x\left( \mathbf{x}_{i}^{v} \right) \right\| ^2+\log\left( {e^  {f_{g}\left(\mathbf{x}_i\right)} +e^ {f_{h,z}\left(\mathbf{x}_i\right)} }\right)-f_{g}\left(\mathbf{x}_i\right) \right]}
\\ &
=\frac{1}{\left| \mathcal{S} _p \right|}\sum_{i=1}^{\left| \mathcal{S} _p \right|}{\left[ \left\| \mathbf{x}_{i}^{v}-f_x\left( \mathbf{x}_{i}^{v} \right) \right\| ^2-f_{g}\left(\mathbf{x}_i\right) \right]}
+\frac{1}{\left| \mathcal{S} _p \right|}\sum_{i=1}^{\left| \mathcal{S} _p \right|}{\log\left( {e^  {f_{g}\left(\mathbf{x}_i\right)} +e^ {f_{h,z}\left(\mathbf{x}_i\right)} }\right)},
\end{aligned}
$$
and
$$\begin{aligned}
\mathcal{L}_p(f) & =\mathbb{E}_\mathbf{x}\left[ \left\| x_{i}^{v}-f_x\left( x_{i}^{v} \right) \right\| ^2-f_{h}\left(\mathbf{x}_i\right) \right]
+\mathbb{E}_\mathbf{x}\left[\log\left( {e^  {f_{g}\left(\mathbf{x}_i\right)} +e^ {f_{h,z}\left(\mathbf{x}_i\right)} }\right)\right].
\end{aligned} $$

Let $\bar{\mathcal{S}} _p$ be the sample set that different from $\mathcal{S} _p$ by only one sample point $ \bar{\mathbf{x}}_{r}$. The empirical risk of the function $f$ on $\bar{\mathcal{S}} _p$ is denoted as $\widehat{\mathcal{L}}^{\prime}_p(f)$. We have

$$\begin{aligned}
& \left|\sup _{f \in \mathcal{F}}| \mathcal{L}_p(f)-\widehat{\mathcal{L}}_p(f)|-\sup _{f \in \mathcal{F}}| \mathcal{L}_p(f)-\widehat{\mathcal{L}}^{\prime}_p(f)|\right| \\
\leq & \sup _{f \in \mathcal{F}}\left|\widehat{\mathcal{L}}_p(f)-\widehat{\mathcal{L}}^{\prime}_p(f)\right| \\
\leq & \sup _{f \in \mathcal{F}}\left|\frac{1}{\left| \mathcal{S} _p \right|}\left(\left\|\mathbf{x}_{r}^{v}-f_x\left(\mathbf{x}_{r}^{v}\right)\right\|^{2}-\left\|\bar{\mathbf{x}}_{r}^{v}-f_x\left(\bar{\mathbf{x}}_{r}^{v}\right)\right\|^{2}
-\left(f_{g}\left(\mathbf{x}_{r}\right)-f_{g}\left(\bar{\mathbf{x}}_{r}\right)\right)\right)\right| \\
& + \sup _{f \in \mathcal{F}}\left|\frac{1}{\left| \mathcal{S} _p \right|}\left(
\log\left( {e^  {f_{g}\left(\mathbf{x}_r\right)} +e^ {f_{h,z}\left(\mathbf{x}_r\right)} }\right)-\log\left( {e^  {f_{g}\left(\bar{\mathbf{x}}_r\right)} +e^ {f_{h,z}\left(\bar{\mathbf{x}}_r\right)} }\right)\right)\right|
\\
\leq & \sup _{f \in \mathcal{F}} \frac{1}{\left| \mathcal{S} _p \right|}\left|\left\|\mathbf{x}_{r}^{v}\right\|^{2}-\left\|\bar{\mathbf{x}}_{r}^{v}\right\|^{2}+\left\|f_x\left(\mathbf{x}_{r}^{v}\right)\right\|^{2}-\left\|f_x\left(\bar{\mathbf{x}}_{r}^{v}\right)\right\|^{2}+2\left\|\mathbf{x}_{r}^{v}\right\|\left\|f_x\left(\mathbf{x}_{r}^{v}\right)\right\|+2\left\|\bar{\mathbf{x}}_{r}^{v}\right\|\left\|f_x\left(\bar{\mathbf{x}}_{r}^{v}\right)\right\|\right|
\\
& + \sup _{f \in \mathcal{F}} \frac{1}{\left| \mathcal{S} _p \right|}\left| f_{g}\left(\mathbf{x}_{r}\right)-f_{g}\left(\bar{\mathbf{x}}_{r}\right)\right|
+ \sup _{f \in \mathcal{F}} \frac{1}{2\left| \mathcal{S} _p \right|}\left|{\left(f_{g}\left(\mathbf{x}_{r}\right)+f_{h,z}\left(\mathbf{x}_{r}\right)\right)-\left(f_{g}\left(\bar{\mathbf{x}}_{r}\right)+f_{h,z}\left(\bar{\mathbf{x}}_{r}\right)\right)} \right|
\\
\leq & \frac{8 D^{2}}{\left| \mathcal{S} _p \right|}.
\end{aligned} $$

Then, we analyze the upper bound of the expectation term $\mathbb{E} \sup _{f \in \mathcal{F}}|\widehat{\mathcal{L}}_p(f)-\mathcal{L}_p(f)|$. Let $\sigma_{1}, \ldots, \sigma_{\left| \mathcal{S} _p \right|}$ be i.i.d. independent random variables taking values in $\{-1,1\}$ and $\bar{\mathcal{S}} _p$ be the independent copy of $\mathcal{S} _p$. We have

$$\begin{aligned}
& \mathbb{E} \sup _{f \in \mathcal{F}}|\widehat{\mathcal{L}}_p(f)-\mathcal{L}_p(f)| \\
\leq & \mathbb{E}_{\mathcal{S} _p, \bar{\mathcal{S}} _p} \sup _{f \in \mathcal{F}}\left|\frac{1}{\left| \mathcal{S} _p \right|}\sum_{i=1}^{\left| \mathcal{S} _p \right|}{\left(\left\|\mathbf{x}_{i}^{v}-f_x\left(\mathbf{x}_{i}^{v}\right)\right\|^{2}-\left\|\bar{\mathbf{x}}_{i}^{v}-f_x\left(\bar{\mathbf{x}}_{i}^{v}\right)\right\|^{2}\right)}\right|
+  \mathbb{E}_{\mathcal{S} _p, \bar{\mathcal{S}} _p} \sup _{f \in \mathcal{F}}\left|\frac{1}{\left| \mathcal{S} _p \right|}\sum_{i=1}^{\left| \mathcal{S} _p \right|}{\left(f_{g}\left(\mathbf{x}_{i}\right)-f_{g}\left(\bar{\mathbf{x}}_{i}\right)\right)}\right|
\\
& +  \mathbb{E}_{\mathcal{S} _p, \bar{\mathcal{S}} _p} \sup _{f \in \mathcal{F}}\left|\frac{1}{\left| \mathcal{S} _p \right|}\sum_{i=1}^{\left| \mathcal{S} _p \right|}{\left(\log\left( {e^  {f_{g}\left(\mathbf{x}_i\right)} +e^ {f_{h,z}\left(\mathbf{x}_i\right)} }\right)-\log\left( {e^  {f_{g}\left(\bar{\mathbf{x}}_i\right)} +e^ {f_{h,z}\left(\bar{\mathbf{x}}_i\right)} }\right)
\right)}\right|
\\
\leq & 2\mathbb{E}_{\mathcal{S} _p, \sigma} \sup _{f \in \mathcal{F}} \left\lvert\, \frac{1}{\left| \mathcal{S} _p \right|}\sum_{i=1}^{\left| \mathcal{S} _p \right|} \sigma_{i}\left\|\mathbf{x}_{i}^{v}-f_x\left(\mathbf{x}_{i}^{v}\right)\right\|^{2}\right|
+ 2 \mathbb{E}_{\mathcal{S} _p, \sigma} \sup _{f \in \mathcal{F}}\left|\frac{1}{\left| \mathcal{S} _p \right|}\sum_{i=1}^{\left| \mathcal{S} _p \right|}{\sigma_{i} f_{g}\left(\mathbf{x}_{i}\right)}\right|
\\
& +  2 \mathbb{E}_{\mathcal{S} _p, \sigma} \sup _{f \in \mathcal{F}}\left|\frac{1}{2\left| \mathcal{S} _p \right|}\sum_{i=1}^{\left| \mathcal{S} _p \right|}{ \sigma_{i}\left({f_{g}\left(\mathbf{x}_i\right) }+{f_{h,z}\left(\mathbf{x}_i\right) }\right)}\right|
\\
\leq & 2 \mathbb{E}_{\mathcal{S} _p, \sigma} \sup _{f \in \mathcal{F}}\left(\frac{1}{\left|\mathcal{S} _p \right|} \sum_{i=1}^{\left| \mathcal{S} _p \right|}\left[\left\|\mathbf{x}_{i}^{v}-f_x\left(\mathbf{x}_{i}^{v}\right)\right\|^{2}\right]^{2}\right)^{\frac{1}{2}}
+ 2 \mathbb{E}_{\mathcal{S} _p, \sigma} \sup _{f \in \mathcal{F}}\left(\frac{1}{\left|\mathcal{S} _p \right|} \sum_{i=1}^{\left| \mathcal{S}_p\right|}\left[f_{g}\left(\mathbf{x}_{i}\right)\right]^{2}\right)^{\frac{1}{2}}
\\
& + 2\mathbb{E}_{\mathcal{S} _p, \sigma} \sup _{f \in \mathcal{F}}\left(\frac{1}{2\left| \mathcal{S} _p \right|} \sum_{i=1}^{\left| \mathcal{S} _p \right|}\left(\left[f_{g}\left(\mathbf{x}_{i}\right)\right]^{2}+\left[f_{h,z}\left(\mathbf{x}_{i}\right)\right]^{2}\right)\right)^{\frac{1}{2}},
\\
\leq & \frac{10 D^{2}}{\sqrt{\left| \mathcal{S} _p \right|}}.
\end{aligned}$$

Thus, according to McDiarmid inequality \cite{mohri2018foundations}, with probability at least $1-\delta$ for any $f \in \mathcal{F}$, we have

$$\mathcal{L}_p(f) \leq \widehat{\mathcal{L}}_p(f)+\frac{10 D^{2}}{\sqrt{\left| \mathcal{S} _p \right|}}+8 D^{2} \sqrt{\frac{\log \frac{1}{\delta}}{2 \left| \mathcal{S} _p \right|}}.$$ 
\end{proof}

Our objective is to collaboratively train all clients to obtain multiple heterogeneous global models capable of handling different view types. This is manifested in the optimization of multiple global objective functions.
For the global objective of handling multiple view types simultaneously, expected risk is defined as $\mathcal{L}_M(f)$, typically optimized in the form of empirical risk minimization, defined as:

$$\widehat{\mathcal{L}}_{M}(f)=\frac{1}{M} \sum_{m=1}^{M} \widehat{\mathcal{L}}_{m}(f).$$
Similarly, for handling individual view types, such as the $v$-th view type, the global objective entails defining the expected risk as $\mathcal{L}_{S^v}(f)$, with empirical risk defined as:

$$\widehat{\mathcal{L}}_{S^v}(f)=\frac{1}{S^v} \sum_{p \in[S^v]} \widehat{\mathcal{L}}_{p}(f)+\frac{1}{M} \sum_{m=1}^{M} \widehat{\mathcal{L}}_{m^v}(f).$$
Now we give the proof of Theorem \ref{th2}.
\begin{proof} \let\qed\relax
According to Lemma \ref{lemma1}, with probability at least $1-\delta$ for any $f \in \mathcal{F}$, we have
    $$\mathcal{L}_M(f) \leq \widehat{\mathcal{L}}_M(f)+\frac{12V D^{2}}{\sqrt{M\left| \mathcal{M} _m \right|}}+9V D^{2} \sqrt{\frac{\log \frac{1}{\delta}}{2 M\left| \mathcal{M} _m \right|}},$$
    where $M$ is the number of multi-view clients and $\left| \mathcal{M} _m \right|$ is the number of samples in each multi-view client. Additionally, we have

    $$\frac{1}{M}\sum_{m=1}^{M}d_{\mathcal{F}}\left(\tilde{\mathcal{D}}_{v}, \tilde{\mathcal{D}}\right)=d_{\mathcal{F}}\left(\tilde{\mathcal{D}}_{v}, \tilde{\mathcal{D}}\right)=\frac{2}{M}\sum_{m=1}^{M} \sup _{f \in \mathcal{F}}\left|\operatorname{Pr}_{\tilde{\mathcal{D}}_{v}}\left[f\right]-\operatorname{Pr}_{\tilde{\mathcal{D}}}\left[f\right]\right|.$$

According to Lemma \ref{lemma2} and Lemma \ref{lemma3}, with probability at least $1-\delta$ for any $f \in \mathcal{F}$, we have

    $$\begin{aligned}
    L_{S^v}(f)\leq & \widehat{L}_{S^v}(f) + \frac{10 D^{2}}{\sqrt{S^v\left| \mathcal{S} _p \right|}}+8 D^{2} \sqrt{\frac{\log \frac{1}{\delta}}{2 S^v\left| \mathcal{S} _p \right|}} \\
    & + \sqrt{\frac{4}{M\left| \mathcal{M} _m \right|}\left(d \log \frac{2 e M\left| \mathcal{M} _m \right|}{d}+\log \frac{4}{\delta}\right)}+d_{\mathcal{F}}\left(\tilde{\mathcal{D}}_{v}, \tilde{\mathcal{D}}\right)+\lambda_{v},
    \end{aligned}$$
 where $S^v$ is the number of single-view clients with $v$-th view type and $\left| \mathcal{S} _p \right|$ is the number of samples in each single-view client.

\end{proof}

\section{Experimental Settings} \label{app:details}
\subsection{Datasets} \label{app:dataset}
We conduct the experiments on the following public datasets. 
\textbf{MNIST-USPS} \cite{peng2019comic} is a widely-used dataset for handwritten digits (0-9) and consists of 5000 examples with two views of digital images. The MNIST feature size is 28 × 28, while the USPS feature size is 16 × 16.
\textbf{BDGP} \cite{cai2012joint} comprises 2500 examples related to drosophila embryos, each represented by a 1750-dimensional visual feature and a 79-dimensional textual feature.
\textbf{Multi-Fashion} \cite{xiao2017fashion} is an image dataset featuring products like Coats, Dresses, and T-shirts, with images sized at 28 × 28. Following the approach in \cite{cui2024novel}, which constructs a three-view version using 30,000 images, each instance includes three different images belonging to the same class. Consequently, the three views of each instance represent the same product with three distinct styles.
\textbf{NUSWIDE} \cite{chua2009nus} is a web image dataset offering multiple views, including 65-dimensional color histogram, 226-dimensional block-wise color moments, 145-dimensional color correlogram, 74-dimensional edge direction histogram, and 129-dimensional wavelet texture. We utilize a total of 5000 samples for the evaluation of our proposed method.
\begin{table*}[!h]
\caption{The statistics of experimental datasets.}
\centering
\label{datasets}
\begin{tabular}{|c|c|c|c|c|c|}
\hline
Dataset    &Clients & Sample & View & Class & Dimension  \\ \hline   
MNIST-USPS & 24   & 5000   & 2 & 10   & {[}[28, 28], [16,16]{]} \\ \hline
BDGP     & 12 & 2500   & 2  & 5  & {[}1750,79{]}    \\ \hline
Multi-Fashion   & 48      & 10000   & 3  & 10   & {[}784, 784, 784{]}           \\ \hline
NUSWIDE & 24   & 5000 & 5  & 5    & {[}65, 226, 145, 74, 129{]} \\ \hline
\end{tabular}
\end{table*}

\subsection{Implementation Details} \label{app:implement}
The models of all methods are implemented on the PyTorch \cite{paszke2019pytorch} platform using NVIDIA RTX-3090 GPUs. For an encoder-decoder pair, the encoder structure follows $\text { Input- } \mathrm{Fc}_{500}-\mathrm{Fc}_{500}-\mathrm{Fc}_{2000}-\mathrm{Fc}_{20}$, and the decoder is symmetric to the encoder. The non-linear mappings $\left\{\mathcal{H}^1\left(\mathbf{Z}^{1}; \Psi^{1}\right),\ldots,\mathcal{H}^V\left(\mathbf{Z}^{V}; \Psi^{V}\right) \right\}$ and $\mathcal{H}\left(\mathbf{Z}; \Psi\right)$ adopt network architectures of $\mathrm{Fc}_{20}-\mathrm{Fc}_{256}-\mathrm{Fc}_{20}$ and $\mathrm{Fc}_{20V}-\mathrm{Fc}_{256}-\mathrm{Fc}_{20}$, respectively. The activation function is ReLU \cite{glorot2011deep}, and the optimizer uses Adam. 
For all the datasets used, the learning rate is fixed at 0.0003, the batch size is set to 256, and the temperature parameters $\tau_m$ and $\tau_p$ are both set to 0.5. Local pre-training is performed for 250 epochs on all datasets. After each communication round between the server and clients, local training is conducted for 10 epochs for the BDGP dataset and 25 epochs for other datasets on each client. The communication rounds between the server and clients are set to $R=5$.
All experiments in the paper involving FMCSC that are not mentioned are performed when $M/S$ = 1:1.

We report some results that number of parameters and runtime by FMCSC to give the reader some information about the computational resources used by the method. Table \ref{tab:runtime} shows that the number of parameters and runtime by FMCSC are small and easy to reproduce.
\begin{table}[!h]
    \caption{Number of parameters and runtime by FMCSC.}
\centering
\label{tab:runtime}
\begin{tabular}{c|cccc}
\hline 
Dataset    &MNIST-USPS & BDGP &Multi-Fashion & NUSWIDE \\ \hline   
Number of parameters per client & 3.4M-6.7M   & 3.5M-7M   & 3.4M-10.1M & 2.8M-13.7M   \\ \hline
Runtime     & 763.8s & 108.5s   & 1183.5s  & 954.7s  \\ \hline 
\end{tabular}
\end{table}

\subsection{Comparison Methods} \label{app:compare}
\begin{figure*}[!h]
      \centering
      \subfigure{\includegraphics[width=0.4\textwidth]{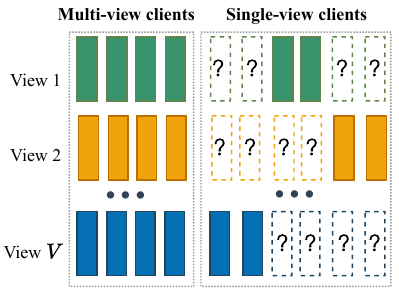}}
      \caption{Comparison strategies.}
      \label{compare}
\end{figure*}

We select 9 state-of-the-art methods, including HCP-IMSC \cite{li2022high}, IMVC-CBG \cite{wang2022highly}, DSIMVC \cite{tang2022deep}, LSIMVC \cite{liu2022localized}, ProImp \cite{li2023incomplete}, JPLTD \cite{lv2023joint}, CPSPAN \cite{jin2023deep}, FedDMVC \cite{chen2023federated} and FCUIF \cite{ren2024novel}. 
Among them, apart from FedDMVC and FCUIF, which are FedMVC methods, all the other comparison methods are centralized incomplete multi-view clustering methods.
To ensure fair comparisons, we simplify the heterogeneous hybrid views scenario in our paper into a hybrid views scenario. Specifically, we concatenate the data distributed among the clients and use them as input for centralized methods, as shown in Figure \ref{compare}.
 Among these, the data from multi-view clients can be considered complete data, while the data from single-view clients can be regarded as missing data.
 It's worth noting that in our reported results, our method operates under the heterogeneous hybrid views scenario, whereas the other comparative methods operate under the hybrid views scenario. Although existing solutions can bypass the challenge of heterogeneous clients by simply concatenating data, the exposure of raw data, due to privacy concerns, may cause more data owners to refuse to participate in collaborative training. In contrast, our method can extract complementary clustering structures across clients without exposing their raw data, offering better privacy protection and performance improvement than current state-of-the-art methods.



\section{Additional Experiment Results} \label{app:add}
\subsection{Convergence Analysis}
\begin{figure*}[!h]
      \centering
      \subfigure[MNIST-USPS]{\includegraphics[width=0.4\textwidth]{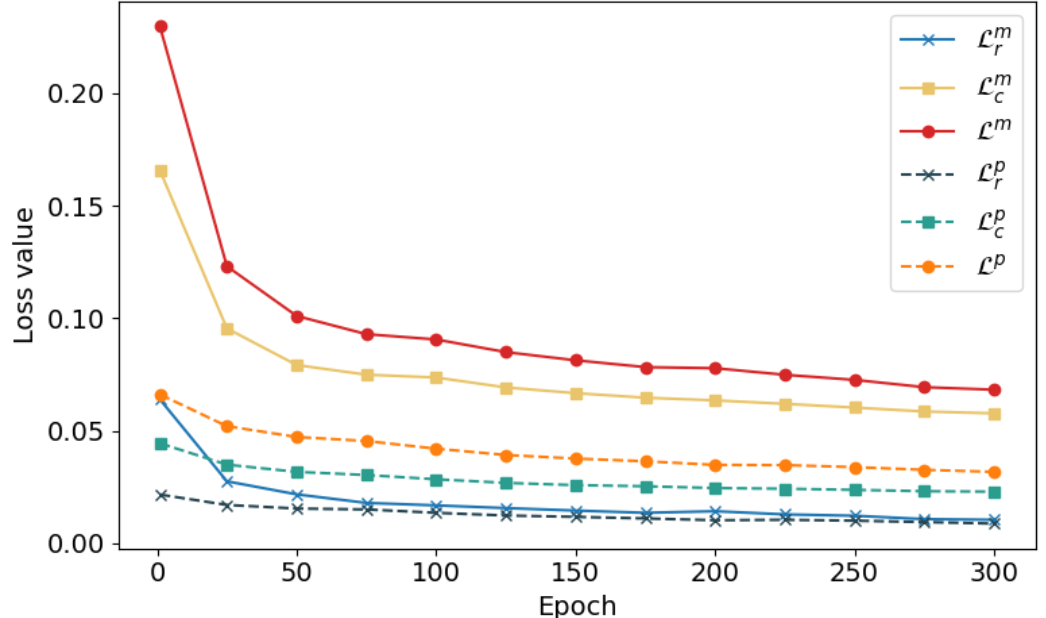}}
      \subfigure[BDGP]{\includegraphics[width=0.4\textwidth]{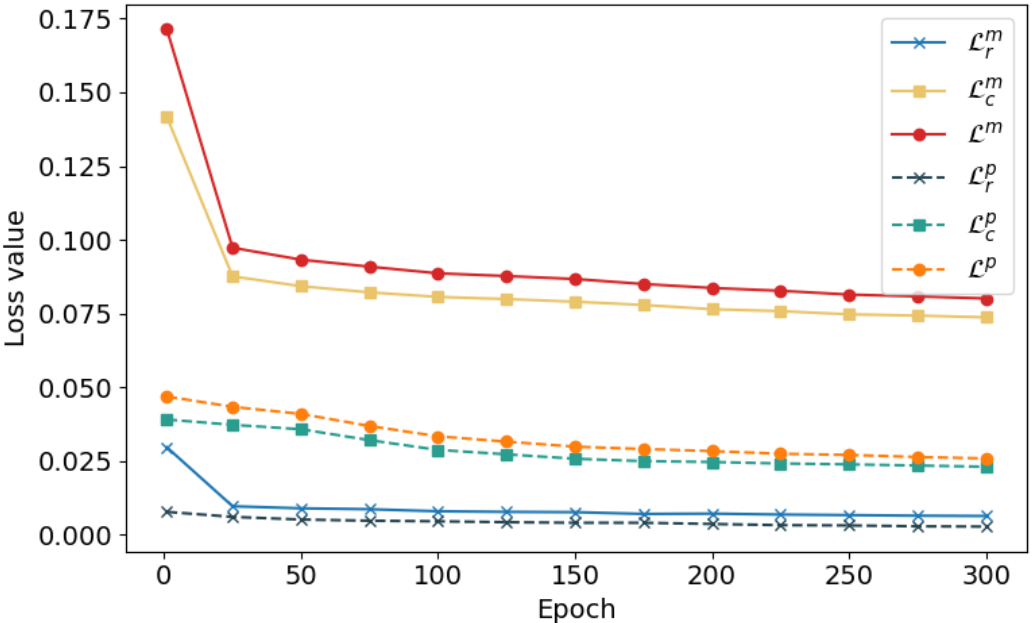}} \\
      \subfigure[Multi-Fashion]{\includegraphics[width=0.4\textwidth]{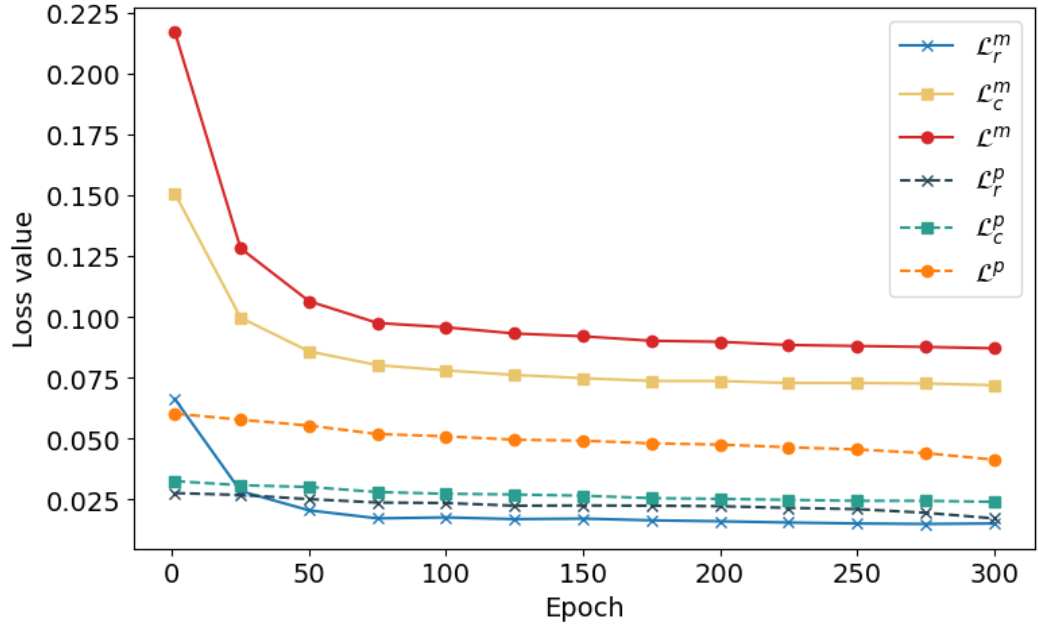}}
      \subfigure[NUSWIDE]{\includegraphics[width=0.4\textwidth]{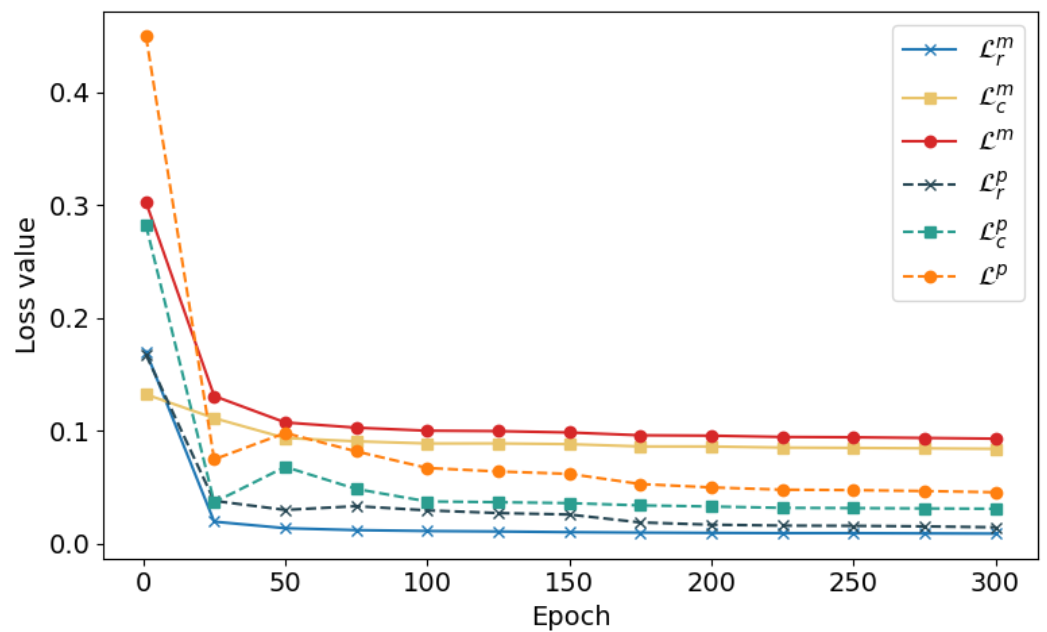}}
      \caption{Convergence analysis on four datasets.}
      \label{loss}
    \end{figure*}

Convergence analysis of the reconstruction loss, consistency loss, and total loss for multi-view clients and single-view clients is conducted using MNIST-USPS, BDGP, Multi-Fashion, and NUSWIDE. 
As illustrated in Figure \ref{loss}, the increasing number of epochs leads to a gradual convergence of all loss functions, ultimately reaching a stable state. This clear observation serves as compelling evidence for the stability and effectiveness of our proposed model.

\subsection{Attributes of Federated Learning}
\begin{figure*}[!h]
      \centering
      \subfigure{\includegraphics[width=0.32\textwidth]{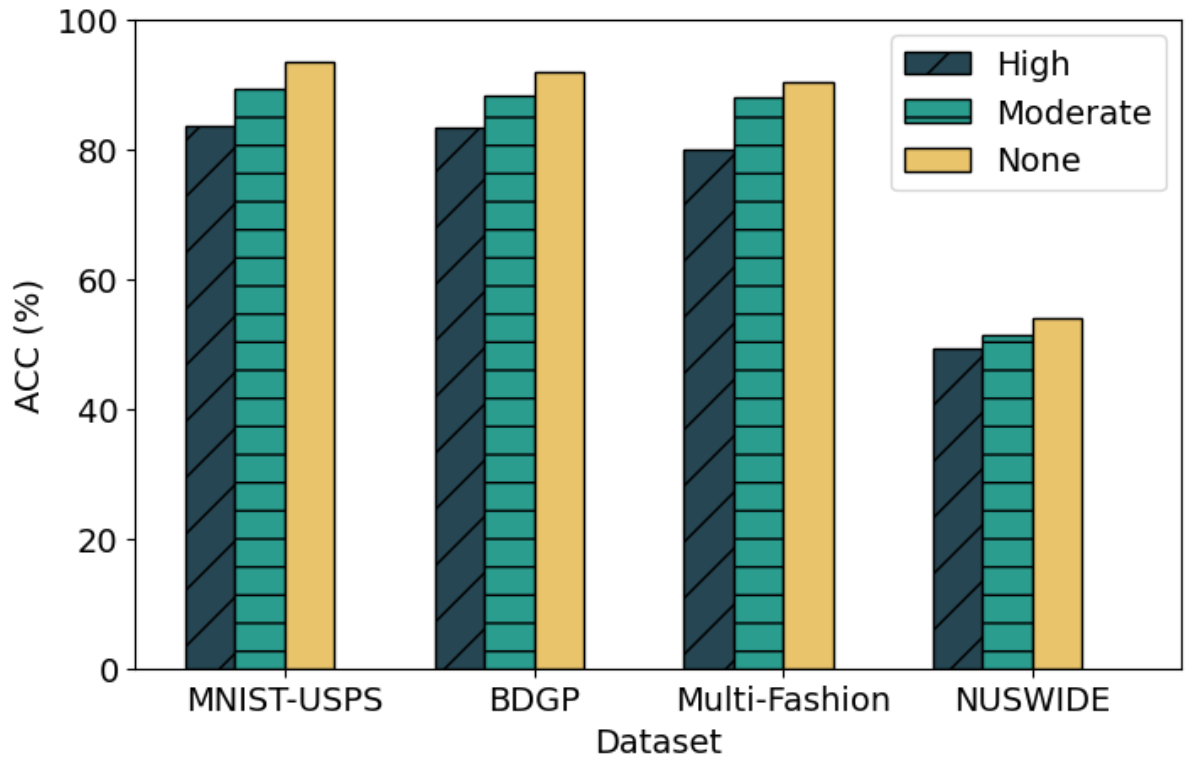}}
      \subfigure{\includegraphics[width=0.32\textwidth]{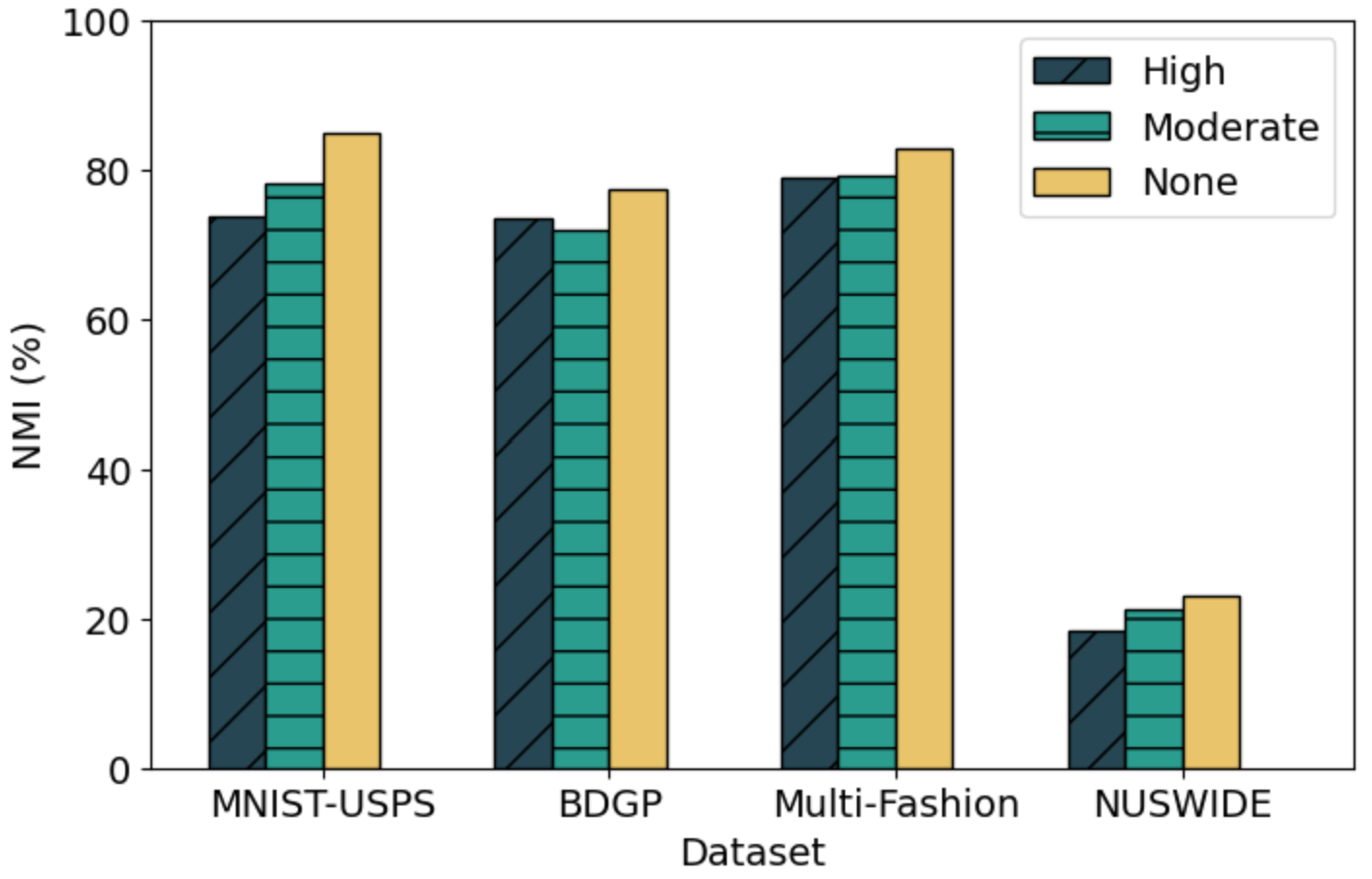}}
      \subfigure{\includegraphics[width=0.32\textwidth]{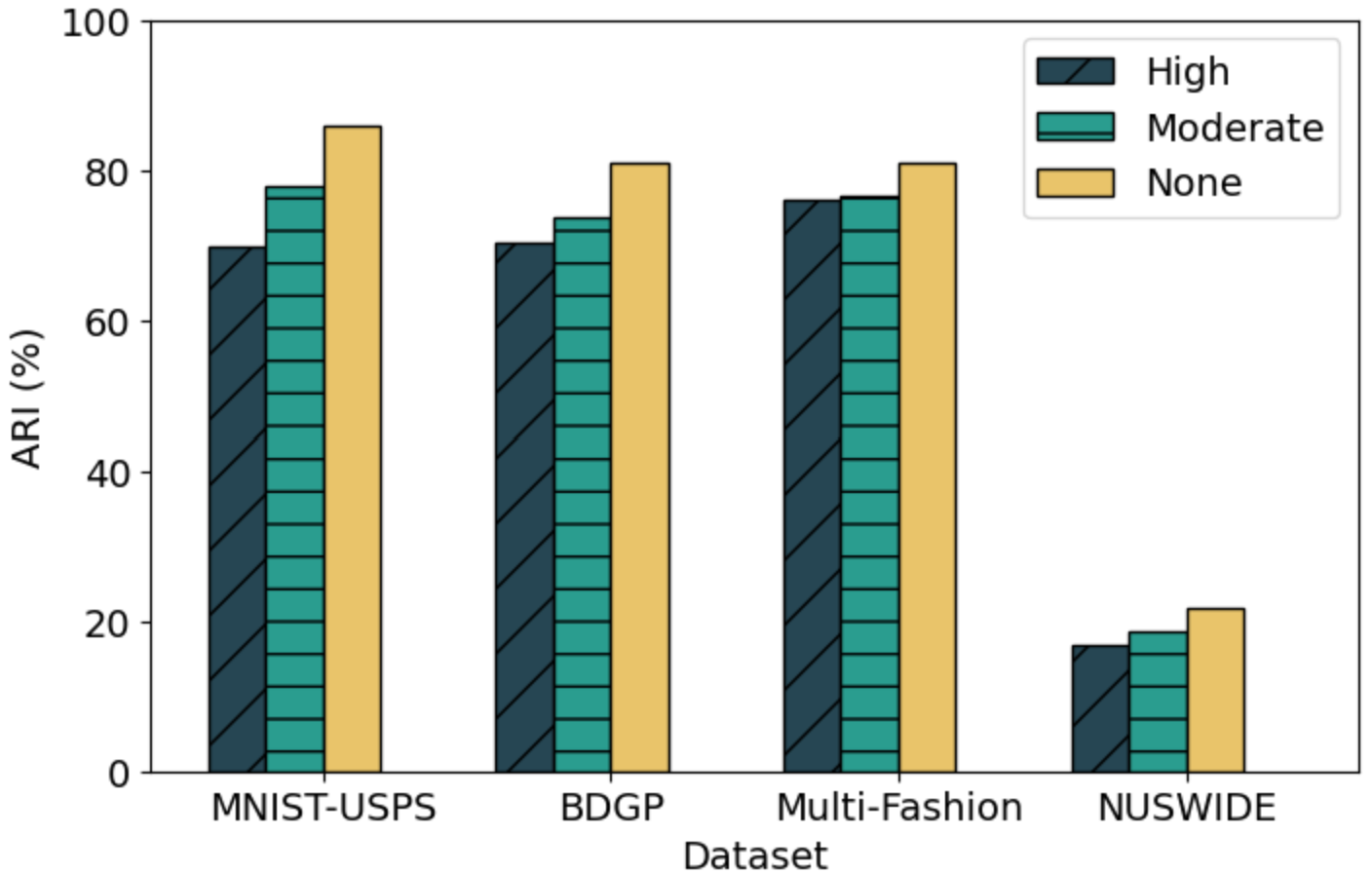}} 
      \caption{Sensitivity to statistical heterogeneity.}
      \label{fl1}
    \end{figure*}

In this section, we present additional experimental results of FMCSC in various federated learning scenarios, including statistical heterogeneity, number of clients, and privacy. 
Figure \ref{fl1} utilizes the Dirichlet distribution Dir($\beta$) to simulate three levels of statistical heterogeneity by setting $\beta$ to 5 (high), 50 (moderate), and 500 (none/IID) on four datasets. 
In addition to reporting additional clustering performance metrics compared to the main text, we observe that FMCSC exhibits adaptability to diverse scenarios, and its performance generally improves as heterogeneity levels decrease. Furthermore, it is noteworthy that on the BDGP dataset, FMCSC demonstrates an increase in NMI values under high heterogeneity, which warrants further exploration.

\begin{figure*}[!h]
      \centering
      \subfigure[MNIST-USPS]{\includegraphics[width=0.32\textwidth]{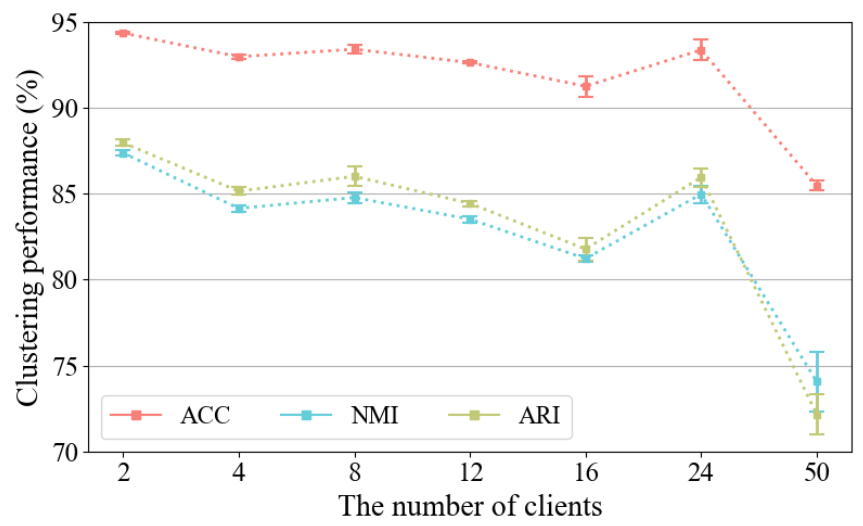}}
      \subfigure[BDGP]{\includegraphics[width=0.32\textwidth]{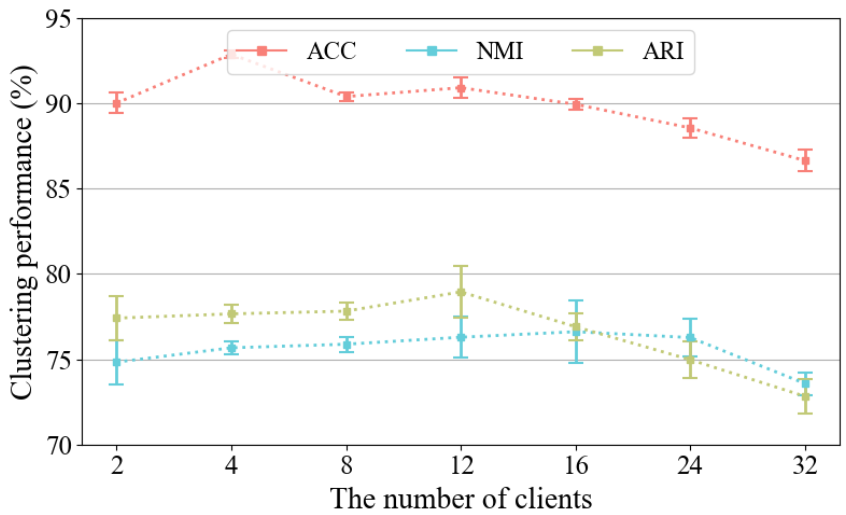}} 
      \subfigure[NUSWIDE]{\includegraphics[width=0.32\textwidth]{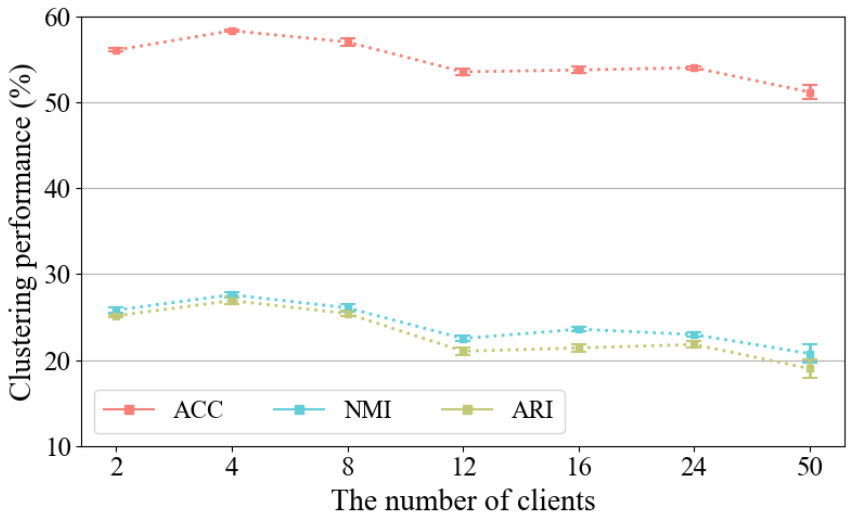}}
      \caption{ Scalability with the number of clients.}
      \label{fl2}
    \end{figure*}

Figure \ref{fl2} presents the impact of the number of clients on clustering performance for MNIST-USPS, BDGP, and NUSWIDE. It is observed that, with an increasing number of clients, the performance of FMCSC shows a slight decline but remains generally stable. 
However, on MNIST-USPS, as the number of clients reaches 50, the clustering performance experiences an unavoidable decrease due to the insufficient number of samples for each client.

     \begin{figure*}[!h]
      \centering
      \subfigure{\includegraphics[width=0.4\textwidth]{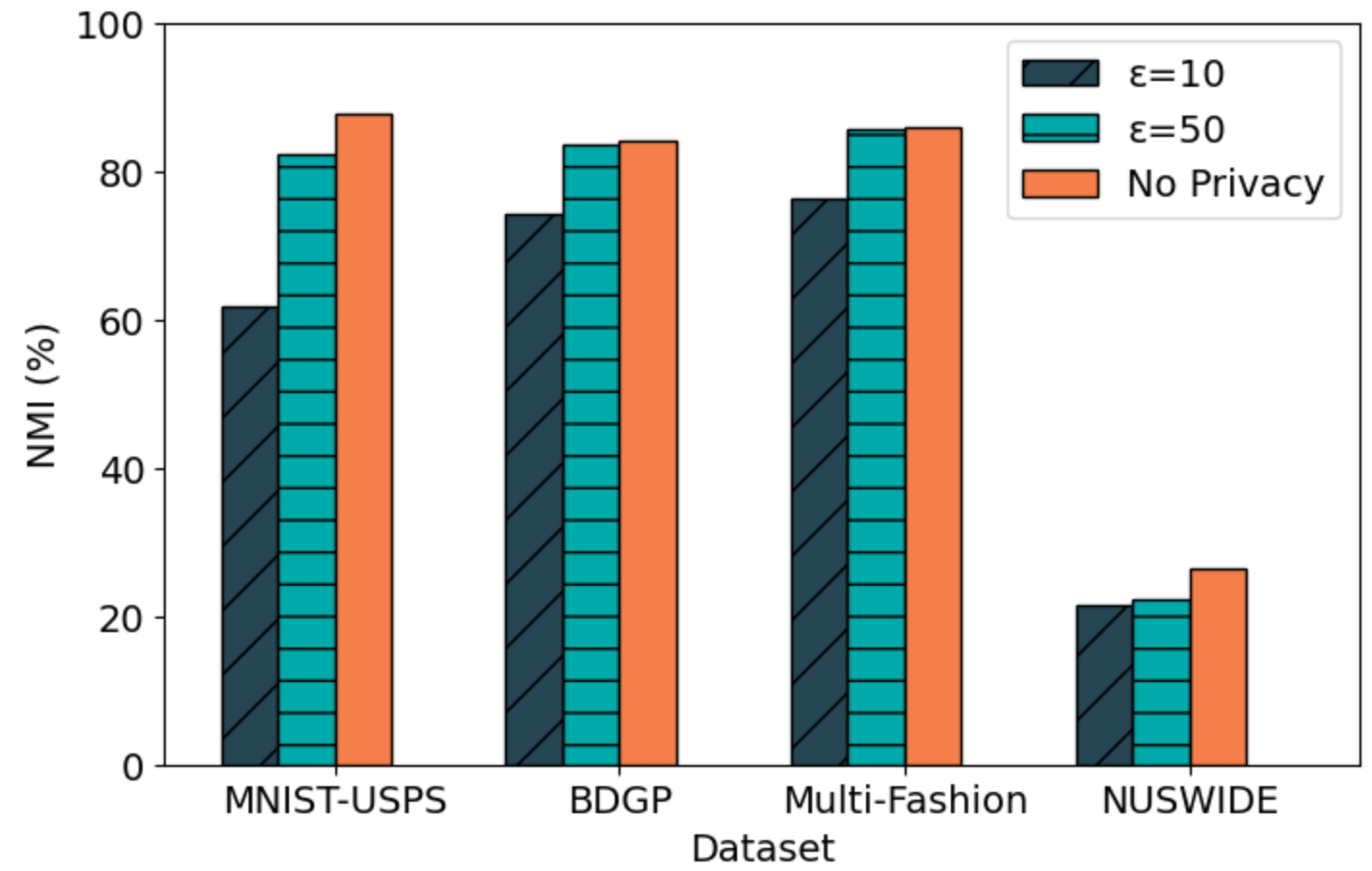}}
      \subfigure{\includegraphics[width=0.4\textwidth]{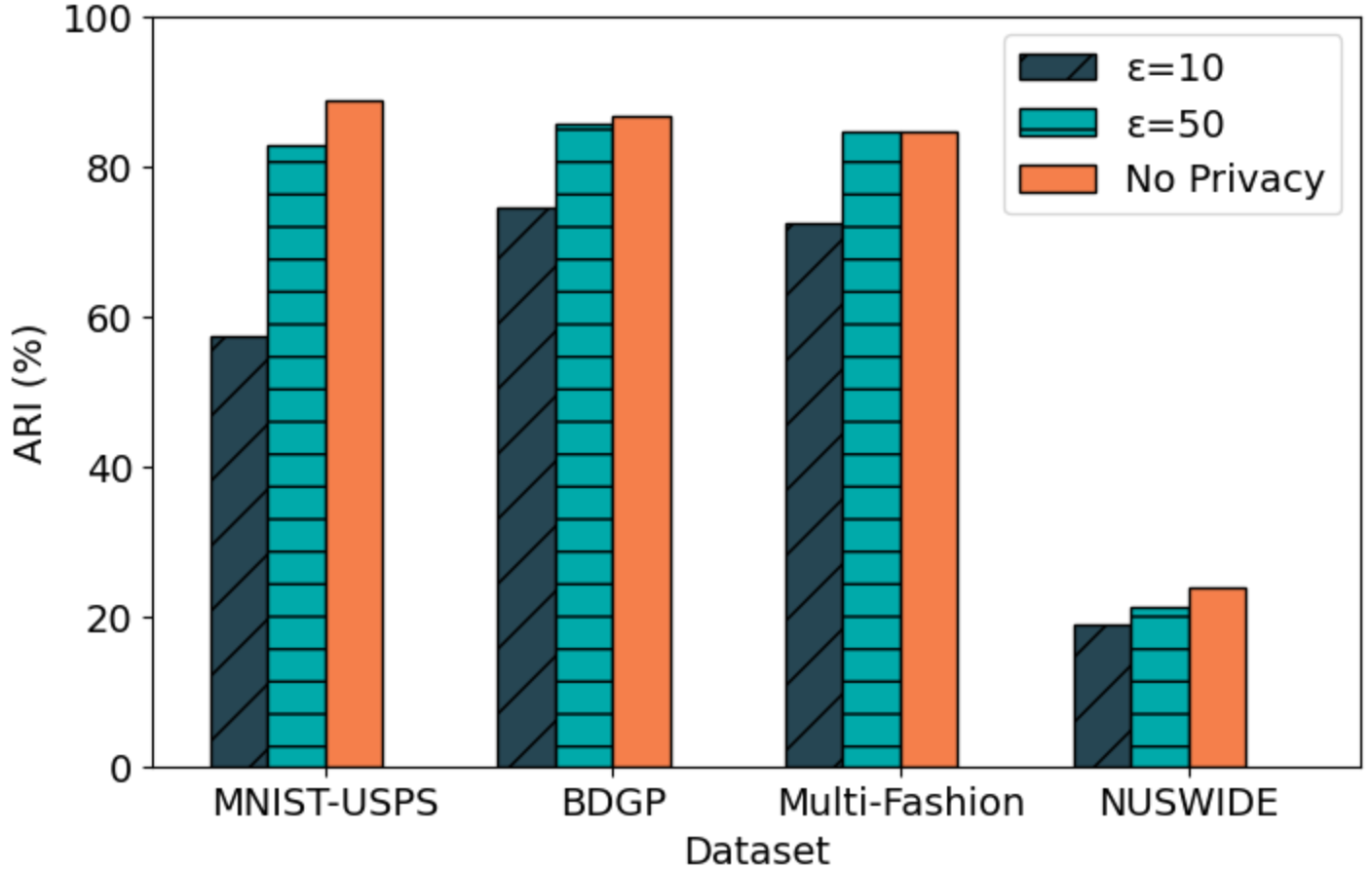}} 
      \caption{Sensitivity under privacy constraints when $M/S$ = 2:1.}
      \label{fl3}
    \end{figure*}

In FMCSC, we assume that all participating parties are semi-honest and do not collude with each other. An attacker faithfully executes the training protocol but may launch privacy attacks to infer the private data of other parties. 
Previous studies have demonstrated that sharing gradients can leak information about raw data \cite{geiping2020inverting}. To address this concern, we utilize differential privacy to further enhance the privacy guarantee of FMCSC. 
Differential privacy algorithms aim to introduce noise during individual data processing to protect against the disclosure of private information \cite{abadi2016deep}. Formally, let $\mathcal{A}$ be a random mechanism that takes dataset $D$ as input and belongs to set $S$. Assuming $D_1$ and $D_2$ are two neighboring datasets differing in only one point, $\mathcal{A}$ is $(\varepsilon, \delta)$ differentially private if:
\begin{equation}
    \operatorname{Pr}\left[\mathcal{A}\left(\mathcal{D}_{1}\right) \in \mathcal{S}\right] \leq \exp (\varepsilon) \cdot \operatorname{Pr}\left[\mathcal{A}\left(\mathcal{D}_{2}\right) \in \mathcal{S}\right]+\delta
\end{equation}
Here, $\varepsilon$ and $\delta$ quantify the individual impact on the overall output in differential privacy. $\varepsilon$ represents the strength of differential privacy, with smaller values indicating stronger privacy protection at the potential cost of increased noise. $\delta \sim \mathcal{O}\left(\frac{1}{\text {number of samples}}\right)$ is used to handle probabilistic exceptional cases.
We implement Laplace noise to achieve differential privacy, and Figure \ref{fl3} reports the clustering performance of FMCSC under different privacy strengths $\varepsilon$.
Our observations reveal that FMCSC achieves both high performance and privacy at $\varepsilon = 50$, especially on the BDGP and Multi-Fashion datasets. However, as the level of noise increases at $\varepsilon = 10$, the performance of FMCSC unavoidably degrades. 

\section{Broader Impacts} \label{app:impact}
Heterogeneous hybrid views are prevalent in real-world scenarios. Our proposed method extends the application domain of existing FedMVC approaches to address complex scenarios such as healthcare and the Internet of Things (IoT). For example, hospitals in metropolitan areas use CT, X-ray, and EHR for disease detection, whereas remote areas often rely on a single detection method. Similarly, smartphones can capture both audio and images simultaneously, while recording devices are limited to collecting audio data only. Furthermore, this research is not expected to introduce any new negative societal impacts beyond those already known.

\section{Limitations} \label{app:lim}
Our model addresses heterogeneous hybrid views in FedMVC, but it idealistically categorizes clients into two types: single-view clients and multi-view clients. In more realistic scenarios, multi-view clients can be further classified into full-view clients and partial-view clients based on the number of view types they have. Such a detailed categorization could encourage more heterogeneous devices to participate in federated learning, thereby enhancing the model's generalization ability and accelerating its application in fields like healthcare and finance. We will continue to explore this problem in future work and apply our findings to real-world scenarios.

\end{document}